\documentclass[a4paper,fleqn]{cas-dc}

\usepackage{amssymb}
\definecolor{hc}{RGB}{0,0,255} % 定义一个自定义颜色
\usepackage{array}
\usepackage{caption}
\usepackage{amsmath}
\usepackage{tabularx}
\usepackage{booktabs}
\usepackage[round,authoryear]{natbib}
\usepackage{subfigure}
\usepackage{algorithm}
\usepackage{adjustbox}
\usepackage{algorithmic}
\usepackage{caption}
\usepackage{booktabs}
\usepackage{soul}
\usepackage{svg}
\usepackage{bm}
\usepackage{color, xcolor}
\usepackage{graphicx}
\usepackage{url}
\usepackage{multirow}
\soulregister\cite7
\soulregister\citep7
\soulregister\ref7

\def\tsc#1{\csdef{#1}{\textsc{\lowercase{#1}}\xspace}}
\tsc{WGM}
\tsc{QE}
\tsc{EP}
\tsc{PMS}
\tsc{BEC}
\tsc{DE}
\begin{document}
	\let\WriteBookmarks\relax
	\def\floatpagepagefraction{1}
	\def\textpagefraction{.001}
	\let\printorcid\relax
	%\pagestyle{empty}
	
	%\shorttitle{Wei~Ai et~al./ Expert Systems with Applications}
	\shortauthors{Wei~Ai et~al.}
	
	\title [mode = title]{Contrastive Multi-graph Learning with Neighbor Hierarchical Sifting for Semi-supervised Text Classification}
	
	
		%\includegraphics[width=1\textwidth]{Fig1.pdf}
		%The overall process of the proposed ConNHS for semi-supervised text classification.
	
		%\item A novel graph-based method for semi-supervised text classification
		%\item A multi-relational text graph based on multiple semantic associations among documents
		%\item A multi-graph learning method consisting of the proposed RW-GCN and CGAN
		%\item A graph contrastive learning approach with neighbor hierarchical sifting strategy

	% \tnotemark[<tnote number>]
	
	
	
	\author[a]{Wei~Ai}
	\ead{aiwei@hnu.edu.cn}	
	\author[a]{Jianbin~Li}
	\ead{jianbinli@csuft.edu.cn}	
	\author[a]{Ze~Wang}
	\ead{zewang@csuft.edu.cn}
	\author[a]{Yingying~Wei}
	\ead{yingying.wei@csuft.edu.cn}	
	\author[a]{Tao~Meng}  %通信作者
	\cormark[1]
	\ead{mengtao@hnu.edu.cn}
		\author[a]{Yuntao~Shou}
	\ead{shouyuntao@stu.xjtu.edu.cn}
	\author[b]{Keqin~Li}
	\ead{lik@newpaltz.edu}
	
	\address[a]{College of Computer and Mathematics, Central South University of Forestry and Technology, Changsha, Hunan 410004, China}
	\address[b]{Department of Computer Science, State University of New York, New Paltz, New York 12561, USA.}
	\cortext[cor1]{Corresponding author}

	%%\author{A}[orcid=0000-0000-0000-0000]
	
	% \fnmark[College of Computer Science and Engineering, Chongqing University of Technology, Chongqing 400054, China]
	%%\ead{1142729000@qq.com}
	% \ead[url]{<URL>}
	% \credit{<Credit authorship details>}
	
	% \affiliation[<aff no>]{organization={},
		% %          citysep={}, % Uncomment if no comma needed between city and postcode
	
	%%\author{B}
	%%\cormark[1]
	%%\ead{xx@xx.com}
	%%\author{C}
	%%\ead{xx@xx.com}
	
	%%\address{College of Computer Science and Engineering, University Name, Chongqing 400054, China}

	%%\cortext[1]{Corresponding author.}
	
	
	
	\begin{abstract}
		Graph contrastive learning has been successfully applied in text classification due to its remarkable ability for self-supervised node representation learning. However, explicit graph augmentations may lead to a loss of semantics in the contrastive views. Secondly, existing methods tend to overlook edge features and the varying significance of node features during multi-graph learning. Moreover, the contrastive loss suffer from false negatives. To address these limitations, we propose a novel method of contrastive multi-graph learning with neighbor hierarchical sifting for semi-supervised text classification, namely ConNHS. Specifically, we exploit core features to form a multi-relational text graph, enhancing semantic connections among texts. By separating text graphs, we provide diverse views for contrastive learning. Our approach ensures optimal preservation of the graph information, minimizing data loss and distortion. Then, we separately execute relation-aware propagation and cross-graph attention propagation, which effectively leverages the varying correlations between nodes and edge features while harmonising the information fusion across graphs. Subsequently, we present the neighbor hierarchical sifting loss (NHS) to refine the negative selection. For one thing, following the homophily assumption, NHS masks first-order neighbors of the anchor and positives from being negatives. For another, NHS excludes the high-order neighbors analogous to the anchor based on their similarities. Consequently, it effectively reduces the occurrence of false negatives, preventing the expansion of the distance between similar samples in the embedding space. Our experiments on ThuCNews, SogouNews, 20 Newsgroups, and Ohsumed datasets achieved 95.86\%, 97.52\%, 87.43\%, and 70.65\%, which demonstrates competitive results in semi-supervised text classification. 
	\end{abstract}

	% Keywords
	\maketitle
	
	%%%%%%%%%%%%%%%%%%%%%%%%
	% \begin{table}[<options>]
		% \begin{tabular*}{\tblwidth}{@{}LL@{}}
			% \toprule
			% \midrule
			%  & \\
			%  & \\
			%  & \\
			%  & \\
			% \bottomrule
			% \end{tabular*}
		% \end{table}
	
	% Use the style of numbering in square brackets.
	% If nothing is used, default style will be taken.
	%\begin{enumerate}[a)]
	%\item
	%\item
	%\item
	%\end{enumerate}
	
	%\begin{itemize}
	%\item
	%\item
	%\item
	%\end{itemize}
	
	%\begin{description}
	%\item[]
	%\item[]
	%\item[]
	%\end{description}

	\section{Introduction}
	\label{sec:introduction}
	Text classification is a crucial task in natural language processing, with a wide range of applications, including sentiment analysis, news categorization, question-answering systems, and spam filtering. Traditional deep learning methods \citep{shi2024robust, lai2015recurrent, tai2015improved, chang2020taming, shou2022conversational, shou2025masked, shou2022object} approach text as a complete whole and capture features from locally continuous word sequences, achieving significant strides. Recent advances in graph-based methods \citep{yao2019graph, linmei2019heterogeneous, yang2022contrastive, piao2022sparse, shou2023comprehensive, shou2024adversarial, meng2024deep} have ushered in a new era of text classification, leveraging the ability of Graph Neural Networks in generating node representations to drive competitive performance.
	
	The first step for graph-based text classification tasks is to break the independence of different data samples by constructing graph topologies for unconnected free texts. The second step involves leveraging the ability of graph neural networks to capture both global and local information to learn text representations. Specifically, existing methods \citep{yao2019graph, lei2021multihop, zhang2020text, lin2021bertgcn, shou2023adversarial, ai2023gcn} treat words and documents as nodes and construct a heterogeneous text graph based on the point-wise mutual information (PMI) relationships between words and the TF-IDF relationships between words and documents. Despite such methods having achieved promising results, they neglect the rich and deep semantics, which is pivotal for capturing the core intent of the text. To account for deep textual semantics, some studies \citep{liu2020tensor, li2021textgtl, ai2024gcn, meng2024multi, shou2024contrastive, shou2024spegcl} propose to construct multi-typed text graphs (i.e., semantic, syntactic, and sequential contexts). TensorGCN \citep{liu2020tensor} executes GCN propagation within different text graphs separately to aggregate neighboring information of nodes. Subsequently, to integrate across-graph features, a virtual graph for nodes at the same positions is constructed to perform inter-graph propagation. TextGTL \citep{li2021textgtl} designs a two-layer parallel GCN to learn document node representations across graphs. Specifically, it independently aggregates information over multiple graphs in the first layer. Then, it performs average pooling on the outputs of the different graphs from the first layer to serve as the input for the second layer. However, these methods have some drawbacks. Firstly, they perform average pooling aggregation on neighboring nodes during intra-graph propagation, neglecting the edge features and the varying relevance between nodes. Secondly, they assign equal weights to different features during the inter-graph propagation, ignoring the intrinsic differences inherent in these features. Overall, the current works neither construct relationships between texts using rich semantics nor propose an effective method for node representation learning across multiple graphs. These shortcomings indicate that exploring a text classification method capable of enhancing semantic connections between texts and improving the multi-graph learning process remains an unresolved challenge.
	
	The emergence of a large amount of unlabeled text has made semi-supervised text classification extremely challenging. Recently, some studies \citep{zhao2023textgcl, sun2022contrastive, li2023graph, shou2024efficient, ying2021prediction, shou2023graph, shou2023czl} have leveraged the self-supervised representation learning capabilities of graph contrastive learning (GCL) to mitigate the issue of label scarcity in text classification. However, these methods rely on explicit graph augmentation to obtain contrastive views. This not only requires prior domain knowledge or trial and error to determine optimal graph augmentation parameters but also may fail to preserve the integrity of task-relevant information through augmentation. Specifically, common augmentation techniques like randomly deleting document nodes or key edges \citep{lan2023contrastive, meng2024masked, shou2024revisiting, ai2024edge} can significantly alter the meaning of the text. This reduces the consistency of learnable representations between contrastive views, thereby misleading the learning process of graph neural networks. Moreover, the fundamental goal of GCL is to design an appropriate contrastive loss function to cluster similar nodes while separating dissimilar nodes. However, current methods like CGA2TC \citep{yang2022contrastive, zhang2024multi, ai2023two, meng2024revisiting, ai2024mcsff} typically employ the NT-Xent contrastive loss function, which is widely used in GCL. Such contrastive loss function considers nodes at the same position as positive samples while treating the remaining nodes within and across views as negative samples. This inevitably leads to selecting document nodes with similar semantics as negative samples and results in similar nodes being far apart in the latent space, which contradicts the fundamental goal of GCL. Existing GCL-based text classification methods result in incomplete information due to their dependence on graph augmentation and produce false negatives on the ground that the use of common contrastive loss. These shortcomings underscore the necessity of developing a novel augmentation-free contrastive learning framework, aimed at overcoming information loss and false negative issues.
	
	To tackle the aforementioned challenges, we propose a novel method of \textbf{Con}trastive multi-graph learning with \textbf{N}eighbor \textbf{H}ierarchical \textbf{S}ifting for semi-supervised text classification, named ConNHS. The proposed method eliminates the need for explicit graph augmentation and introduces a novel contrastive loss function to optimize representation learning. Firstly, we extract titles, keywords, and events to construct a multi-relational text graph that can represent more latent semantic connections. Secondly, to avoid the loss of structural information caused by graph augmentation, we separate the multi-relational text graph to derive semantic subgraphs (corresponding to titles, keywords, and events). This provides multiple views for the graph contrastive learning stage. Subsequently, we propose a relation-aware graph convolutional network (RW-GCN) to perform intra-graph propagation within each semantic subgraph, which considers the varying correlations between document nodes and incorporates edge feature information. Moreover, considering the differences among semantic subgraphs, we design a cross-graph attention network (CGAN) for inter-graph propagation to obtain fused node representations, effectively harmonizing the feature information from different subgraphs. Additionally, we present the neighbor hierarchical sifting loss (NHS) to circumvent the false negative pairs that could undermine contrastive learning efforts. Specifically, NHS masks the first-order neighbors of the anchor, as the construction of the multi-relational text graph is dependent on the homophily assumption, i.e., connected document nodes tend to share the same label. Furthermore, NHS draws signals from the similarity score matrix of the fused node representations, excluding high-order neighbors with high similarity to the anchor from being chosen as negatives. This dual approach, rooted in graph structure and node attributes, prevents similar nodes from being distanced in the latent space. Finally, we input the fused node representations obtained from multiple subgraphs into a logistic regression classifier to achieve the final classification results.
	
	The main contributions of this article can be summarized as follows:
	
	\begin{itemize}
		\item
		We harness core features to forge a multi-relational text graph that contains multiple semantic connections among documents. Meanwhile, we propose RW-GCN to leverage edge features and capture varying correlations between nodes. We also design CGAN to coordinate the fusion of feature information across graphs.
		
		\item
		We propose a contrastive learning method for semi-supervised text classification that does not require graph augmentation. Our innovative contrastive loss function effectively optimizes negative selection and avoids the occurrence of false negatives, thus providing clearer clustering boundaries for downstream text classification.
		
		\item
		We test the proposed method on four real-world datasets (including Thucnews, Sogounews, 20NG, and Ohsumed), and the results demonstrate the effectiveness of ConNHS for semi-supervised text classification tasks.
	\end{itemize}
	
	The rest of this paper is organized as follows: Section \ref{sec:relatedwork} introduces related work, Section \ref{sec:proposedmethod} presents the detailed method, Section \ref{sec:experiments} gives the experimental setup and results, and finally, Section \ref{sec:conclusion} gives a brief conclusion.

	%\hl{The contributions of this paper are summarized as follows: 1) Text text text text text text text text text text text text text text text text text text text text text text text text text.} 2) Text text text text text text text text text text text text text text text text text text text text text text text text text. 3) Text text text text text text text text text text text text text text text text text text text text text text text text text.
	%================:2：Related work===================%
	\section{Related Work}
	\label{sec:relatedwork}
	\subsection{Deep Learning for Text Classification}
	In the early stages of text classification, methods primarily focused on machine learning-based techniques, heavily relying on feature engineering dependent on specific domain knowledge and experience. With the advent of deep learning models, the need for feature engineering has significantly been alleviated, as these models possess the capability to learn textual features automatically. Specifically, TextCNN \citep{kim2014convolutional} is the first attempt to transfer the CNN model, widely applied in computer vision, to text classification tasks. It extracts local features using multiple filters, but this method struggles to capture long-range dependencies in text sequences effectively. RNN-based methods, such as TopicRNN \citep{dieng2016topicrnn} and RNN-Capsule \citep{wang2018sentiment}, can address the long-term dependency problem and learn more comprehensive text representations, but they may encounter issues like gradient explosion or vanishing gradients. In recent years, Transformer-based pre-trained models \citep{kenton2019bert, shi2024robust, liu2019roberta, shou2023graphunet, ai2024graph, shou2024low, shou2024graph}, with their exceptional semantic understanding capabilities, have been widely applied in text classification tasks, achieving significant results. Despite the remarkable success of sequence-based deep learning models in text classification tasks, they still exhibit certain inherent limitations. For example, they primarily focus on token-level information processing, potentially overlooking the complex intertextual relationships and higher-level semantic structures. These limitations highlight the necessity of exploring text classification methods with enhanced semantic understanding capabilities to better address complex classification challenges.

	\subsection{GNN for Text Classification}
	Graph Neural Networks (GNN) are deep learning models designed for graph-structured data. They are widely used in fields such as social network analysis, recommendation systems, and molecular chemistry. GNNs leverage the information from nodes and edges in a graph to effectively represent and learn from graph data. The Graph Convolutional Network (GCN) \citep{kipf2016semi} model achieves good results by performing spectral convolutions on node features, making it widely applicable for tasks like node classification and graph embedding. The Graph Attention Network (GAT) \citep{velivckovic2018graph} introduces an attention mechanism that allows the model to assign different weights when aggregating information from neighboring nodes, enhancing the model's expressive power and flexibility. With the rapid development of graph neural networks, a variety of graph-based text classification models have also emerged.
	
	These models can be broadly categorized into document-level and corpus-level types. Document-level methods treat words as nodes and construct an independent text graph for each document, effectively mining contextually relevant word relationships. For example, Text-Level-GNN \citep{huang2019text} uses a sliding window approach, employing a limited number of nodes and edges in each text graph to reduce memory and computational overhead. Meanwhile, TextFCG \citep{wang2023text} builds a single graph for all words in a text, marking edges with various contextual relationships, and adopts GNN and GRU for text classification. On the other hand, corpus-level methods capture the global structural information of a corpus by constructing one or more graphs containing both word and document nodes, which include various relationships like word-word and word-document. TextGCN \citep{yao2019graph} constructs the entire corpus as a heterogeneous graph, using word nodes as intermediaries for information transfer and facilitating inter-document information exchange through a two-layer GCN. TensorGCN \citep{liu2020tensor} constructs a text graph tensor to capture semantic, syntactic, and sequential contextual information and uses both intra-graph and inter-graph propagation to harmonize heterogeneous information from multiple graphs. However, these methods often fall short in fully capturing textual semantic information when constructing graph structures, leading to an inadequate understanding of the deeper meanings within the text. Additionally, when employing multi-type text graphs, these approaches face challenges in learning both intra and inter graphs due to feature discrepancies between different types of nodes. This inconsistency in features can hinder the model to accurately grasp global semantic relationships and effectively propagate information.
	
	\subsection{Graph Contrastive Learning}
	Graph contrastive learning \citep{xu2021infogcl, mo2022simple, xia2022simgrace, yang2022dual} is a technique for extracting features efficiently using unlabeled data. Its core idea is to generate positive and negative samples by transforming the original data, thereby reducing the distance between similar data and increasing the distance between dissimilar data in the feature space, achieving a clustering-like effect. The process of graph contrastive learning mainly covers three key stages. The first is the graph data augmentation stage, which is crucial to ensure the difference and diversity between views and has a significant impact on the final model’s performance. Second is embedding learning, which involves encoding node samples to generate contrastive samples. Finally, the calculation of contrastive loss includes defining positive and negative sample pairs, thereby promoting the model to learn more discriminative node features.
	
	Recently, in the field of graph contrastive learning, numerous efficient methods and applications have gradually emerged. For instance, MVGRL \citep{hassani2020contrastive} employs graph diffusion techniques for graph-level data augmentation on the original input graph, thereby obtaining views containing richer global information. GraphCL \citep{you2020graph}, on the other hand, explores various graph augmentation strategies to address the heterogeneity issue in graph data. Simultaneously, GCA \citep{zhu2021graph} introduces an adaptive data augmentation scheme, moving away from the traditional practice of uniformly dropping edges or perturbing features. Instead, it emphasizes the enhancement of essential nodes and edges and the disruption of node features to obtain more effective views. GCNSS \citep{miao2022negative} effectively mitigates the false negative pairs problem in graph contrastive learning by utilizing label information. Additionally, NCLA \citep{shen2023neighbor} proposes a new learnable graph augmentation strategy, generating higher-quality contrastive views. For GCL-based text classification methods, ConKGNN \citep{lan2023contrastive} constructs a unified graph that includes text and related knowledge graph (KG) information and introduces contrastive learning to accomplish the text classification task. However, the random graph augmentation it utilizes can lead to unpredictable information loss. TextGCL \citep{zhao2023textgcl} simultaneously trains GCN and BERT, utilizing contrastive learning loss to learn precise text representations. It lacks a discerning mechanism in the selection of negative samples, inevitably introducing false negatives.
	
	%=================3：Proposed method===================%
	\section{Proposed method}         
	\label{sec:proposedmethod}
\begin{figure*}
		\centering
		\makebox[\textwidth][c]{%
			\includegraphics[width=1\textwidth]{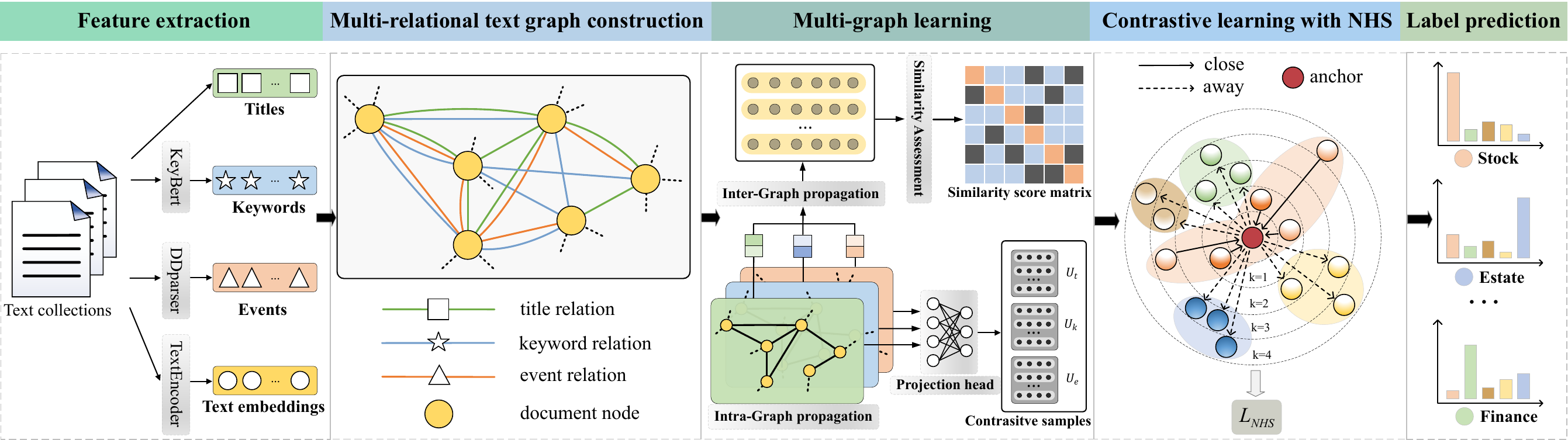}
		}
		\caption{Flow chart of the proposed ConNHS. Initially, we construct a multi-relational text graph by leveraging inherent core features (titles, keywords, events) to establish semantic connections among texts while encoding textual content as initial node representations. Subsequently, relational separation yields distinct subgraphs, upon which intra-graph and inter-graph propagation are performed to obtain contrastive samples and similarity score matrix. During Contrastive learning with NHS, negative selection is optimized to encourage more explicit cluster boundaries (minimizing intra-class distances while maximizing inter-class distances; distinct colors indicate different clusters). Ultimately, predicted labels are assigned to document nodes via a logical classifier.
		}
		\label{Fig1}
	\end{figure*}

	In this section, we first provide a brief overview of our proposed ConNHS method, followed by a detailed explanation of its constituent modules. The overall process of ConNHS is illustrated in Figure \ref{Fig1}. As shown, our proposed ConNHS comprises five main stages: (1) Feature extraction: For semantically enriched texts, we start from the semantic level by extracting the titles, keywords, and events of the texts. These core features are used as the basis for constructing the text graph. (2) Multi-relational text graph construction: Inspired by the intrinsic logic that humans use to classify texts, we construct multiple document-to-document relationships by calculating the similarity of core features in the embedding space. The constructed text graph contains more latent semantic connections between document nodes. (3) Multi-graph learning: To avoid explicit graph augmentation, we separate the multi-relational text graph into different semantic subgraphs. We propose a relation-aware graph convolutional network to perform intra-graph propagation within each subgraph. This method fully considers edge features and the varying correlations between nodes, thus aggregating more significant neighborhood information. Additionally, given the differences in node features across subgraphs, we design a cross-graph attention network. It facilitates inter-graph propagation to obtain the fused text representations, thereby enabling a comprehensive and nuanced understanding of textual congruence. (4) Contrastive learning with NHS: To acquire precise text representations, we apply a novel graph contrastive learning methodology for model training. By presenting an innovative contrastive loss to refine negative selection, the ubiquitous quandary of false negatives in GCL is substantially mitigated, thereby enhancing the fidelity and robustness of the text representations. (5) Label prediction: We leverage the pre-trained model to obtain fused text representation and input them into a logistic regression classifier to predict the label of each text. In the subsequent sections of the paper, we will describe each component of the ConNHS model in detail.

	%==============Feature extraction
	\subsection{Feature extraction}
	The fundamental logical judgment for humans to ascertain the domain of a text is recognizing the features that can represent the core intention of the text. For instance, when the word "goalkeeper" is present in two pieces of news, our cognitive systems are inclined to categorize both texts under the sports domain. The inclination is rooted in the understanding that texts sharing similar snippets of information or vocabulary are likely to emanate from the same sphere. Drawing inspiration from this human-centric logic for classifying texts, we aim to extract various core features to forge links between texts that are otherwise unconnected.
	
	\textbf{Title:} Serving as the introductory sentence of an article, titles typically encapsulate information pertinent to the topic of text, providing a high-level synopsis of the content. Fundamentally, the title is constructed as the first sentence imbued with comprehensive semantic information, necessitating no further processing. The titles set can be formalized as $Title = \{t_1, t_2,\cdots, t_n\} $ , where $t_i$ is the title of the text $i$.
	
	\textbf{Event:} Moreover, we incorporate the concept of events to achieve a more sophisticated level of textual representation. Typically, an event is characterized as an action or condition that has transpired or is currently happening. Utilizing events as a means of text representation is a widely acknowledged approach, offering a clearer conveyance of textual information than mere sentences or phrases. Therefore, considering that events are mainly composed of objects and the actions they emit, we introduce the definition of event \citep{zhang2022robust} as follows:
	\begin{equation}
		\label{event}
		Event = (W, C, O),
	\end{equation}
	where $W$ represents the action that occurs during the event, $C$ is the factor that causes the event to happen, and $O$ is another object that is involved in the event. The main task of event extraction is identifying and extracting the subject, action, and object. We choose DDparser \citep{zhang2020practical} and Stanza \citep{zhang2021biomedical} as extraction tools to extract events from Chinese and English texts, respectively. The events can be formalized as $EventSet = \{E_1, E_2,\cdots, E_n\} $, where $E_i$ is the set of events extracted from the text $i$.
	
	\textbf{Keyword:} Events distil the essence of a text under the assumption that its semantic core is anchored in specific paragraphs or sentences. Nonetheless, in instances where the content is more scattered, particularly in lengthier texts, the event-centric approach might fall short of capturing textual semantics at the granularity of individual words. Consequently, to address this granularity gap and ensure a comprehensive understanding of the textual thematic breadth, we establish semantic relationships between texts based on keywords at a more fine-grained level. We choose KeyBert\footnote{\url{https://github.com/MaartenGr/KeyBERT}} as the extraction tool, and the extracted keywords are formalized as $KeywordSet = \{K_1, K_2,\cdots, K_n\} $, where $K_i$ is the set of keywords extracted from the text $i$.
	
	The core features delineated above and text contents are transmuted into a computable format via a text embedding model, with the preeminent choice being models pre-trained on extensive corpora. The preference is rooted in two fundamental advantages: firstly, pre-trained models are imbued with a robust knowledge base, endowing them with superior semantic comprehension capabilities; secondly, these models exhibit context sensitivity, which is crucial for adeptly navigating the complexities of homographs—words identical in spelling but divergent in meaning. With the aim of precisely representing the core features and textual contents, this study will employ a text encoder that is comprised of the LangChain\footnote{\url{https://github.com/langchain-ai/langchain}} framework and BGE-M3 \citep{chen2024bge}, a variant version of Bert. This combination is tasked with converting each title, keyword, and event into vector representations, which are instrumental in constructing a multi-relational text graph and laying the groundwork for intricate semantic relationships between texts.

	%==============Text graph construction
	
	\subsection{Multi-relational text graph construction}
	A common graph construction strategy for graph-based text classification methods \citep{yao2019graph} involves analyzing the PMI relationships between words and the TF-IDF relationships between words and documents. This approach, however, overlooks the deep semantic information that can represent the underlying relationships within the text. Therefore, we calculate the semantic similarity between the extracted features to facilitate the construction of multiple semantic relationships between document nodes, corresponding to title relationships, keyword relationships, and event relationships. Based on the rich features inherent in the text, the constructed text graph can maximize the connections between similar documents. Formally, considering the multi-relational text graph as: $G=\{V,A,R\}$ contains document nodes and relationship collection, where $V=\{v_1,v_2,…,v_n\}$, $v_i$ represents the document node $i$, and $n$ represents the number of document nodes. Moreover, $A$ is the adjacency matrix of the text graph. The edge sets are represented by $R=\{T, K, E\}$, corresponding to the title, keyword, and event relationship.
	
	\textbf{Node representation:} The majority of texts are interspersed with information unrelated to the main topic, underscoring the necessity for meticulous preprocessing of text content within the source space. For example, the primary body of news articles often encompasses author signatures, names of news agencies, and additional elements that are tangential to the core intention of the article. The process of obtaining the initial node representation is as follows:
	\begin{equation}
		\label{text-encoder}
		m_i = TextEncoder(c_i),
	\end{equation}
	where $c_i$ is the preprocessed content of text $i$, $m_i \in \mathbb{R}^d$, and $d$ is the dimension of node representation.
	
	\textbf{Title relation:} Titles serve as succinct summaries of textual content and are pivotal in the classification of texts. It is observed that texts belonging to the same category often exhibit a notable similarity in their titles. To capitalize on this observation, we introduce a scoring mechanism designed to quantify the similarity between titles. The quantification of the semantic similarity between titles $t_i$ and $t_j$ can be expressed in the following manner:
	\begin{equation}
		\label{title-simi-score}
		S_{ij}^{t} = Sim(t_i, t_j),
	\end{equation}
	where $Sim(\cdot,\cdot)$ denotes the cosine similarity measure, which quantifies the magnitude of the angle formed by two vector representations $X$ and $Y$ in the latent space. It can be formulated as:
	\begin{equation}
		\label{cosine}
		Sim(X, Y) = \frac{\sum_{i=1}^{n}(x_i \cdot y_i)}{(\sum_{i=1}^{n} x_i^2)^\frac{1}{2} \cdot (\sum_{i=1}^{n} y_i^2)^\frac{1}{2}},
	\end{equation}
	for the title relation between text $i$ and text $j$, we define it as follows:
	\begin{equation}
		\label{title-relation}
		R_{ij}^{t} = \begin{cases}
			1 & \text{if } S_{ij}^{t} >  \rho_t,
			\\
			0 & \text{otherwise},
		\end{cases}
	\end{equation}
	if the quantified semantic similarity $S_{ij}^t$ between titles $t_i$ and $t_j$ transcend the predefined threshold $\rho_t$, the title relation $R_{ij}^t$ shall be established.

	\textbf{Event relation:} Events describe the core intent of a document and thus can serve as a significant feature in constructing the potential connections of texts. Different documents often contain multiple events. Two events sharing a similarity score exceeding the pre-determined threshold $\rho_e$ are considered as a matching event pair. For text $i$ and text $j$, we quantify the relatedness of their constituent events as follows:
	\begin{equation}
		\label{event-batch-simi-score}
		L_{ij}^{e} = \{(e_a, e_b) | e_a \in E_i, e_b \in E_j, Sim(e_a, e_b) > \rho_e\},
	\end{equation}
	where $L_{ij}^{e}$ is the list of matching event pair.

	\begin{equation}
		\label{event-relation}
		R_{ij}^{e} = \begin{cases}
			1 & \text{if } Counter(L_{ij}^{e}) >  \gamma_e,
			\\
			0 & \text{otherwise},
		\end{cases}
	\end{equation}
	where $Counter(\cdot)$ is a utility function that serves to count the elements within a list. If the matching event pairs shared by text $i$ and text $j$ exceed the minimum association coefficient $\gamma_e$, the event relation $R_{ij}^{e}$ will be established.
	
	\textbf{Keyword relation:} Keywords are vital in understanding the theme of a text, offering a new perspective for establishing semantic relations between text nodes. The keywords exhibiting a similarity score that surpasses the predetermined threshold $\rho_k$ are deemed to be a matching keyword pair. The procedure for establishing keyword relation bears a resemblance to that for event relation. It can be formulated as follows:
	\begin{equation}
		\label{keyword-batch-simi-score}
		L_{ij}^{k} = \{(k_a, k_b) | k_a \in K_i, k_b \in K_j, Sim(k_a, k_b) > \rho_k\},
	\end{equation}
	where $L_{ij}^{k}$ is the list of matching keyword pair.
	\begin{equation}
		\label{keyword-relation}
		R_{ij}^{k} = \begin{cases}
			1 & \text{if } Counter(L_{ij}^{k}) >  \gamma_k,
			\\
			0 & \text{otherwise},
		\end{cases}
	\end{equation}
	if the number of matching keyword pairs is greater than the minimum association coefficient $\gamma_k$, the keyword relation $R_{ij}^{k}$ shall be instantiated.
	
	Titles, keywords, and events serve as foundational elements in constructing connections between texts, each offering a unique perspective on the features that define their semantic relationships. The multifaceted approach enables texts that are potentially analogous to share information across their respective nodes, thereby facilitating a more enriched and nuanced text representation learning.
	
	\subsection{Multi-graph learning}
	Recent studies have proposed constructing multi-typed text graphs for text classification tasks, but they have limitations during multi-graph learning. Firstly, they discount the edge features and use average pooling to aggregate neighborhood information during the intra-graph propagation. This aggregation method assumes that all neighboring document nodes are equally important, disregarding the diversity of documents. Secondly, they overlook the differences in node features across different text graphs during inter-graph propagation.
	
	To maintain the integrity of task-relevant graph structural information while providing diverse views for graph contrastive learning, a crucial step is separating the multi-relational text graph according to the relationship type, leading to the creation of semantic subgraphs, as illustrated in Figure \ref{Fig1}. After that, we perform intra-graph and inter-graph propagation on these derived semantic subgraphs.
	
	\textbf{Intra-graph propagation:} Rather than conventional GCN \citep{kipf2016semi}, we propose a relation-aware graph convolution network which consists of a relation-aware aggregation operator $g(\cdot;\theta_{g})$ and a transformation operator $f(\cdot;\theta_{f})$. In detail, let $x_i$ denote the feature representation of node $v_i$ at the $l$-th layer, the aggregation operation can be formally expressed as follows:
	\begin{equation}
		\label{relation-aware-aggregation}
		g(\cdot;\theta_{g})	= \sum_{x_{j}^{l}\in\mathcal{N}(x_{i}^{l})} h(x_{j}^{l}-x_{j}^{l};\theta_{h}) \cdot (x_{j}^{l}-x_{j}^{l}),
	\end{equation}
	where $h(\cdot;\theta_h)$ represents a learnable function parameterized by $\theta_h$, whose purpose is to ascertain the important weights quantifying the correlation between document nodes. The instantiation of $h(\cdot;\theta_h)$ is achieved through a fully connected layer followed by a sigmoid activation. Let $\mathcal{N}(x_i^l)$ denote the feature set of neighboring nodes $x_i^l$ at the $l$-th layer, wherein $x_j^l$ corresponds to the feature representation of the neighbor node $v_j$. Notably, the edges $x_{j}^{l}\in\mathcal{N}(x_{i}^{l})$,  $(x_j^l - x_i^l)$ connecting the centroid node and its neighboring nodes serve as input to the aggregation operator. In other words, $h(x_j^l - x_i^l;\theta_h)$ can be interpreted as the importance weights characterizing the relation between $x_j^l$ and $x_i^l$. Furthermore, we aggregate all the weighted correlation edge features as the aggregating features in the graph, consequently capturing the latent relations among diverse document nodes. Concerning the transformation operator $f(\cdot;\theta_f)$, we concatenate the node feature $x_i^l$ with the aggregating features obtained from $g(\cdot;\theta_g)$ as its input. The updated feature $x_i^{l+1}$ of node $v_i$ by the RW-GCN at the $(l+1)^{th}$ layer can be formally defined as follows:
	\begin{equation}
		\label{relation-aware-transformation}
		x_{i}^{l+1} = f([x_{i}^{l}, g(\cdot;\theta_{g})]; \theta_{f}),
	\end{equation}
	where $x_{i}^{l+1}\in\mathbb{R}^{2 \times d}$, $[\cdot,\cdot]$ is a concatenation operation. $\theta_{f}$ is the independent learnable weight matrix to transform the input features.

	Within the realm of graph contrastive learning, a conventional strategy involves augmenting the input graph to generate two distinct views, followed by the extraction of feature representations from these views using a graph encoder. This methodology, reliant on graph augmentation, presents two primary challenges: Firstly, prevalent graph augmentation techniques, such as edge dropping and attribute masking, risk compromising the structural integrity and semantic content of the graph. For example, the elimination of critical edges could adversely affect the learning of node representations. Secondly, the application of graph augmentation techniques typically necessitates iterative fine-tuning to identify optimal parameters, a process that can be both time-consuming and imprecise. In response to these issues, our method obviates the necessity for intricate graph augmentation procedures. Instead, we employ relation-aware GCN to process semantic subgraphs which inherently possess distinct adjacencies. This approach enables the derivation of varied and diverse views without the introduction of graph augmentations, thereby preserving the original graph structural and semantic integrity.

	\textbf{Intra-graph propagation:} After intra-graph propagation, each document node learns unique feature information under different semantic relationships. Therefore, we design a cross-graph attention network to coordinate and integrate diverse feature information. The process of aggregating document node representations from different subgraphs can be formalized as follows:
	
	\begin{equation}
		\label{cross-graph-attention-score}
		\alpha_{r} = softmax(k^{T}tanh(p(x_{i, r};\theta_{p}))),
	\end{equation}
	where $x_{i,r}$ is the representation of the document node $v_i$ at the relation $r$ subgraph, and $\alpha_{r}$ is the attention weight. $p(\cdot;\theta_{p})$ is a feedforward neural network parameterized by $\theta_{p}$. Then, we use the computed attention weights to perform cross-graph information propagation. The process is as follows:
	\begin{equation}
		\label{cross-graph-attention-aggre}
		H^{\prime}_{i} = (\otimes\{\alpha_{r} \cdot x_{i, r}\} \vert_{r=1}^{R}),
	\end{equation}
	where $\otimes$ is the sum operator, and $R=\{title, keyword, event\}$ is the set of semantic relations. $H^{\prime}_{i}$ is the fused representation of the document node $v_{i}$.

	\textbf{Projection mapping:} To mitigate the impact of irrelevant features across contrasting views while preserving the most salient information, certain graph contrastive learning approaches advocate mapping node embeddings onto a specific latent space. Consistent with prior approaches, we employ a projection head $q(\cdot;\theta_{q})$ to transform the node embedding representation into a tailored latent space prior to computing the contrastive loss objective. In this paper, the mapping process can be formulated as follows:
	\begin{equation}
		\label{MLP}
		u_i = q(h_i; \theta_{q}),
	\end{equation}
	where $h_i$ is the learned representation of document node $v_i$, and $u_i$ is the mapping result, which will be regarded as contrastive node samples for contrastive learning.

	%==============Contrastive learning with NHS
	
	\subsection{Contrastive learning with NHS}
	A key step in graph contrastive learning is designing an appropriate contrastive loss function to cluster similar nodes while separating dissimilar nodes. In traditional contrastive learning paradigms, the contrastive loss NT-Xent typically selects nodes at corresponding positions across views as positive samples for the anchor node. Conversely, all remaining nodes, irrespective of their positioning within or across views, are designated as negative samples. However, such negative selection that NT-Xent adopts inevitably induces false negative pairs. It inadvertently broadens the gap between nodes that are inherently similar, thereby contravening the foundational goal of GCL.

	The fundamental principle underlying contrastive learning can be generally encapsulated as employing a transformation function $f(\cdot)$ to map the input node representation $x$ onto $f(x)$, such that the resultant mapping adheres to the following inequality constraint:
	\begin{equation}
		\label{GCL-distance}
		Dist(f(x_i), f(x_i^+)) \ll Dist(f(x_i), f(x_i^-)),
	\end{equation}
	where $x_i^+$ denotes a positive sample exhibiting similarity to $x_i$, while $x_i^-$ represents a negative sample dissimilar to $x_i$. The function $Dist(\cdot, \cdot)$ serves as a similarity measure employed to quantify the degree of similarity between the node embedding representations.

	\begin{figure*}
		\centering
		\includegraphics[width=0.85 \textwidth]{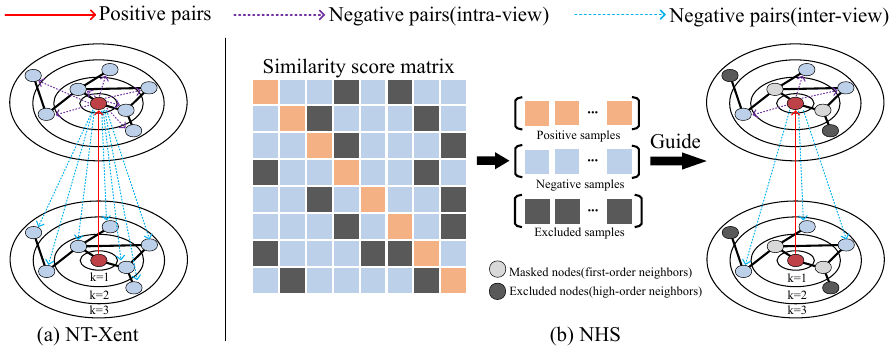}
		\caption{Definition of negative pairs in different contrastive losses. Figure \ref{Fig2} showcases different negative selection definition strategies. Specifically, both NT-Xent and NHS recognize nodes positioned identically across views as positive samples for the anchor. However, NT-Xent designates all remaining nodes as negatives. In contrast, NHS masks first-order neighbors of the anchor document node and the positive nodes based on the graph homophily principle, and also, based on the similarity score matrix of fused node representations, as shown in (b), it excludes those high-order neighbors that exhibit significant similarity to the anchor. To facilitate a more straightforward interpretation, sifted hierarchical neighbors that will not be included in the contrastive learning process are indicated with specific colors in (b). }
		\label{Fig2}
	\end{figure*}

	\textbf{Neighbor Hierarchical Sifting:}
	To address this challenge, our work proposes the neighbor hierarchical sifting loss designed to prevent the incidence of false negative pairs generation, as illustrated in Figure \ref{Fig2}. In keeping with the conventional loss for identifying positive pairs, we continue to regard nodes situated in matching positions across views as positive samples relative to each other. Importantly, extending our consideration to the graph homophily, not only the first-order neighbors of the anchor node but also positive nodes across different views are masked and removed from the negatives. Furthermore, for high-order neighboring nodes that belong to the same category yet lack direct connections, their selection as negative samples can also compromise contrastive learning efficacy. To address this, we access the similarity score matrix between document nodes and identify those high-order neighbors exhibiting substantial similarity to the anchor node, excluding them from negative sample selection. The presented neighbor hierarchical sifting loss significantly mitigates potential false negatives by accounting for the characteristics of neighbors across different hierarchical levels, thereby improving the contrastive learning process and enhancing the quality of learned node representations.
	
	\textbf{Contrastive loss:} Based on the proposed negative selection strategy, we present a novel graph contrastive loss function neighbor hierarchical sifting loss (NHS). In this paper, node $i$ in view $r^{\prime}$ is selected as the anchor node, its embedding is expressed as $u_i^{(r^{\prime})}$, and its contrastive loss can be formulated as follows:
	\begin{equation}
		\label{loss-overall}
		\mathcal{L}(u_i^{(r^{\prime})}) = -log \frac{\xi_{inter}^{pos}}{\xi_{inter}^{pos} + \xi_{intra}^{neg} + \xi_{inter}^{neg}},
	\end{equation}
	the different terms in the above equation can be broken down into:
	\begin{equation}
		\label{loss-pos-inter}
		\xi_{inter}^{pos} = (\otimes{\{e^{Dist(u_i^{(r^{\prime})}, u_i^{(r)})/\tau}\} \vert_{r=1}^{R}}),
	\end{equation}
	\begin{equation}
		\label{loss-neg-intra}
		\xi_{intra}^{neg} = \sum\nolimits_{v_j \subset D_i^{(r^{\prime})}}^{} (e^{Dist(u_i^{(r^{\prime})}, u_j^{(r^{\prime})})/\tau}),
	\end{equation}
	\begin{equation}
		\label{loss-neg-inter}
		\xi_{inter}^{neg} = (\otimes{\{\sum\nolimits_{v_j \subset D_i^{(r)}}^{} e^{Dist(u_i^{(r^{\prime})}, u_j^{(r)})/\tau}\} \vert_{r=1}^{R}}),
	\end{equation}
	where $R=\{title, keyword, event\}, r^{\prime} \notin R$, $\otimes$ is the sum operator. $u_i^{(r)}$ is the representations of node $i$ at the same position in view $r$. And $D_i^{(r)}$ is the negative sets of node $i$ from view $r$. Specifically, the function $Dist(\cdot, \cdot)$ is instantiated as the cosine similarity measure, which quantifies the degree of similarity between two vector representations. The final contrastive loss NHS, defined as averaged over all nodes among the three views, is computed as follows:
	\begin{equation}
		\label{final-loss}
		\mathcal{L}_{NHS} = \frac{ {(\otimes \{\sum^{N}_{i=1} \mathcal{L}(u_i^{(r)})\} \vert_{r=1}^{R} )} } {q \cdot N},
	\end{equation}
	where $R=\{title, keyword, event\}$, $q=|R|$, $\otimes$ is the sum operator, and $N$ is the number of node in contrastive view.

	%==============Label prediction
	
	\subsection{Label prediction}
	In the evaluation phase, we use the pre-trained RW-GCN and CGAN models to obtain text representations for the test data. For the text $i$, its final text representation is denoted as $\mathcal{T}_i$. Then, $\mathcal{T}_i$ will be input into a logistic regression classifier to obtain the classification results:
	\begin{equation}
		\label{lr-classifer}
		p_i = LRClassifier(\mathcal{T}_i).
	\end{equation}
	where $p_i$ denotes the predicted label of text $i$.
	
	To sum up, the ConNHS can be summarized as Algorithm \ref{alg:connhs}:
	
	\begin{algorithm}
		\caption{The overall process of ConNHS}
		\label{alg:connhs}
		\begin{algorithmic}[1]
			\REQUIRE A text corpus \bm{$C$}, similarity threshold \bm{$\rho_t$}, \bm{$\rho_k$}, \bm{$\rho_e$}, minimum association coefficient \bm{$\gamma_k$},  \bm{$\gamma_e$}.
			\STATE \bm{$titles$}, \bm{$keywords$}, \bm{$events$} = FeatureExtraction(\bm{$C$})
			\STATE \bm{$G$} = \bm{$(V, A, R)$} = BuildGraph(\bm{$titles$}, \bm{$keywords$}, \bm{$events$}, \bm{$\rho_t$}, \bm{$\rho_k$}, \bm{$\rho_e$}, \bm{$\gamma_k$}, \bm{$\gamma_e$})
			\STATE \bm{$\hat{G_t}$}, \bm{$\hat{G_k}$}, \bm{$\hat{G_e}$} = Separation(\bm{$G$})
			\FOR{$t = 1$ to $T$}
			\STATE \bm{${H_t}$}, \bm{${H_k}$}, \bm{${H_e}$} = \bm{$RW-GCN(\hat{G}_t, \hat{G}_k, \hat{G}_e)$};
			\STATE \bm{$H^{\prime}$} = \bm{$CGAN$} \bm{$({H_t}, {H_k}, {H_e})$};
			\STATE \bm{${U_t}$}, \bm{${U_k}$}, \bm{${U_e}$} = \bm{$Mapping({H_t}, {H_k}, {H_e})$};
			\STATE \bm{$score$} = SimilarityAssessment\bm{$(H^{\prime})$};
			\STATE \bm{$negatives$} = NHS\bm{$(V, A, score)$}; /*Negative selection by NHS*/
			\STATE Compute contrastive loss \bm{$\mathcal{L}_{NHS}$} with the refined \bm{$negatives$} via Eq.(\ref{loss-overall}) and Eq.(\ref{final-loss});
			\STATE Update parameters by applying gradient descent minimize \bm{$\mathcal{L}_{NHS}$}.
			\ENDFOR
			\STATE Get the text representations $\mathcal{T}$ via the pre-trained RW-GCN and CGAN.
			\STATE Predict the labels of $\mathcal{T}$ via the logistic regression classifier.
			\RETURN The predicted labels of each document node.
		\end{algorithmic}
	\end{algorithm}
	
	%================:4：Experiments===================%
	\section{Experiments}
	\label{sec:experiments}
	In this section, we select four common text classification datasets and verify the effectiveness of our proposed method. Next, we will introduce the datasets and preprocessing, comparison methods, experimental settings, evaluation indicators, experimental results, and experimental result analysis.

	\subsection{Datasets and preprocessing}
	We select three news topic classification datasets (including two Chinese and one English news dataset) and a dataset in the medical field. A brief introduction to the dataset is as follows:

	\textbf{ThuCNews}\footnote{\url{http://thuctc.thunlp.org/}}: The ThuCNews corpus constitutes a news document collection derived through filtering the historical data from the Sina News RSS subscription channel spanning the period of 2005 to 2011, encompassing 14 news categories and comprising approximately 830,000 news articles. Considering the device factor and balancing the dataset, we randomly sample 5000 entries in each of the 14 categories.
	
	\textbf{SogouNews}\footnote{\url{https://huggingface.co/datasets/sogou_news}}: The SogouNews Corpus, furnished by SogouLabs, represents a news dataset encompassing SogouCA and SogouCS, comprising approximately 27,000 news items distributed across ten distinct categories. To attain a balanced distribution within the dataset, around 3000 samples were randomly sampled from each category, with the constraint that the character count of each sample exceeded 500.
	
	\textbf{20NG}\footnote{\url{http://qwone.com/~jason/20Newsgroups/}}: The 20 News Corpus is an English text classification dataset containing newsgroup posts in 20 categories. There are 18,846 articles in total, with an average of about 1,000 articles per category.
	
	\textbf{Ohsumed}\footnote{\url{https://disi.unitn.it/moschitti/corpora.htm}}: The Ohsumed corpus is derived from the MEDLINE database, a bibliographic repository of significant medical literature curated by the National Library of Medicine. It encompasses 23 categories and a total of 7,400 articles. Given that each article is annotated with one or more tags, the highest-level tag is selected as the definitive label for the experimental setting.
	
	\begin{table}
		\caption{Summary statistics of the benchmark dataset.}
		\renewcommand\arraystretch{0.8}
		\setlength\tabcolsep{1mm}%调列距
		\centering
		\resizebox{1\linewidth}{!}{
			\begin{tabular}{@{}c|ccccc@{}}
				\toprule[1.3pt]
				& \#Docs &  \#Train &  \#Test & \#Classes & \#Avg.Length \\ \midrule[1.3pt]
				ThuCNews & 84,000 & 67,200 & 16,800 & 14 & 539.75       \\  \cmidrule(r){1-1}
				SogouNews  & 30,000 & 24,000 & 6,000 & 10 & 502.4        \\ \cmidrule(r){1-1}
				20NG   & 18,846 & 15,076 & 3,770 & 20 & 221.26            \\ \cmidrule(r){1-1}
				Ohsumed   & 7,400 & 5,920 & 1,480 & 23   & 135.82         \\ \bottomrule[1.3pt]
		\end{tabular}}
		\label{table1}
	\end{table}

	\textbf{Pre-processing}: First, we filtered two Chinese news data sets according to the length of the text. The average length of the filtered news exceeded 500, which can verify the effectiveness of our proposed method in classifying longer texts. Secondly, noisy information unrelated to text category labels, such as the name of the author of the article and publication time, were removed from all datasets. Finally, Table \ref{table1} lists the summary statistics of the benchmark datasets.
	
	\subsection{Comparison of methods}
	In order to verify the effectiveness of our proposed method, we compared the four datasets mentioned above with the following eight state-of-the-art methods, which are:
	
	\textbf{PV-DBOW} \citep{le2014distributed}: It is a paragraph vector model and ignores the word order in the text. Logistic regression is used as a classifier.
	
	\textbf{fastText} \citep{joulin2017bag}: The approach utilizes the mean of word/n-gram embeddings to represent document embeddings, subsequently feeding these aggregated vectors into a linear classifier for further analysis.
	
	\textbf{TextCNN} \citep{kim2014convolutional}: It is a type of traditional deep learning model and harnesses convolutional layers to autonomously and adaptively extract spatial hierarchies of features from the input data, thereby enabling the model to discern intricate patterns and relationships within the text.
	
	\textbf{RNN-Capsule} \citep{wang2018sentiment}: This model employs a capsule network-enhanced Recurrent Neural Network (RNN) for conducting sentiment analysis.
	
	\textbf{Bi-LSTM} \citep{yao2019graph}: A variant of the LSTM model is commonly used in text classification tasks.
	
	\textbf{Bert-large} \citep{sun2024text}: It is a pre-trained language model based on the Transformer architecture. Based on the well-trained model, it is used for downstream text classification tasks after fine-tuning.
	
	\textbf{TextGCN} \citep{yao2019graph}: It is a model that uses graph convolutional neural networks for text classification. By building a graph structure and utilizing the representation learning capabilities of graph neural networks, it can effectively capture the semantic relationships between texts and improve the accuracy of text classification.
	
	\textbf{HAN} \citep{wang2019heterogeneous}: It proposes a novel dual-layer attention mechanism, encompassing node-level attention and semantic-level attention. Node-level attention is employed to quantify the salience of the relation between the centroid node and its heterogeneous neighboring nodes, whereas semantic-level attention serves to ascertain the relative importance of distinct meta-paths.
	
	\textbf{RGCN} \citep{schlichtkrull2018modeling}: It handles different types of nodes and relationship edges through relationship-specific graph convolution layers and node representation layers and obtains rich semantic information by iteratively updating node representations.

	\textbf{TGNCL} \citep{li2023graph}: It builds a graph for each document and develops a contrastive learning regularization to learn fine-grained word representations.
	
	\subsection{Experiment setting and evaluation criteria}
	In this section, we present the specific details of the experiment. Before conducting many experiments, the text will be preprocessed to remove irrelevant noise information. The second step is to extract the core feature information of the text (including keywords and events). Correspondingly, the title is a complete semantic sentence that can be obtained directly without special processing. For document nodes, which represent each text in this framework, the embedding representation encodes the text attributes using the LangChain framework, and the pre-trained embedding model BGE-M3 is used as the initial representation of the node. It is worth noting that our experimental results are the averages of 10 runs with different weight initializations.
	
	During the training process, if the training loss does not decrease for more than 50 consecutive epochs, the model is deemed to have reached convergence. Our method uses the Adam optimizer in the deep learning framework Pytorch. The training and testing processes of all datasets were completed on a computer equipped with Intel core i9-12900k CPU and Nvidia Geforce RTX3090.

	We choose $Accuracy$, $Precision$, and $F1$ scores, common indicators in text classification tasks, to measure the effectiveness of our proposed method. $Accuracy$ represents the proportion of correctly classified samples to the total number of samples. $Precision$ indicates the proportion of correctly classified positive samples among all samples classified as positive. $F1$ takes into account precision and recall, making the evaluation more comprehensive.They can be formulated as:
	
	%\textcolor{hc}{\textbf{Accuracy}: }
	\begin{equation}
		\label{Accuracy}
		Accuracy = \frac{TP + TN}{TP + TN + FP + FN}
	\end{equation}
	
	%\textcolor{hc}{\textbf{Precision}: }
	\begin{equation}
		\label{Precision}
		Precision = \frac{TP}{TP + FP}
	\end{equation}
	
	%\textcolor{hc}{\textbf{F1}: }
	\begin{equation}
		\label{F1}
		F1 = \frac{2PR}{P + R}
	\end{equation}
	
	where $TP$ (True Positives) represents the number of samples correctly classified as category $Y_i$. $FP$ (False Positives) refers to the number of samples from other categories incorrectly classified as $Y_i$. $TN$ (True Negatives) indicates the number of samples from other categories correctly classified as not $Y_i$. $FN$ (False Negatives) are the samples belonging to category $Y_i$ but incorrectly classified into other categories. Additionally, $R$ stands for Recall, which is the proportion of correctly predicted positive samples out of all actual positive samples.

	\subsection{Experiment results and analysis}
	\subsubsection{Performance on text classification}
	
	Table \ref{table2} delineates the accuracy, precision, and F1 scores achieved by various methodologies across four benchmark datasets. Predominantly, the proposed ConNHS outperforms the baseline methods, showcasing its superior text classification prowess. The proposed ConNHS achieved accuracy improvements of 1.12, 0.30, 1.51, and 2.12 on the ThuCNews, SogouNews, 20NG, and Ohsumed datasets, respectively. We observed that the improvements of ConNHS on English datasets were more pronounced compared to the Chinese news datasets. This can be attributed to the fact that the baseline methods already achieved accuracy rates exceeding 90\% on the Chinese news datasets. For Precision and F1 scores, ConNHS is likewise the most competitive method, consistently ranking among the top across all datasets. It is worth noting that RGCN demonstrated outstanding performance on the ThuCNews dataset, achieving the best Precision. Additionally, Bert-large and TGNLCL exhibited remarkable stability, with no significant performance fluctuations across multiple datasets. In contrast, the PV-DBOW model performed poorly in terms of precision and F1 score, lacking the competitiveness compared to deep learning models.

	\captionsetup[table]{skip=0pt}
	\begin{table*}[h]
		\renewcommand\arraystretch{1.3}
		\setlength{\tabcolsep}{5pt}
		\caption{Test accuracy(\%), P(\%), and F1 score(\%) for different models on two Chinese datasets and two English datasets.}
		\begin{center}
			\resizebox{\linewidth}{!}{
				\begin{tabular}{ccccccccccccccccc}
					\toprule[1.5pt]
					\multirow{2}{*}{\textbf{Method}}&&\multicolumn{3}{c}{\textbf{ThuCNews}}&&\multicolumn{3}{c}{\textbf{SogouNews}}&&\multicolumn{3}{c}{\textbf{20NG}}&&\multicolumn{3}{c}{\textbf{Ohsumed}} \\
					\cline{3-5} \cline{7-9} \cline{11-13} \cline{15-17}
					~ && \textbf{\textit{Acc}}& \textbf{\textit{P}}& \textbf{\textit{F1}} && \textbf{\textit{Acc}}& \textbf{\textit{P}}& \textbf{\textit{F1}} && \textbf{\textit{Acc}}& \textbf{\textit{P}}& \textbf{\textit{F1}} && \textbf{\textit{Acc}}& \textbf{\textit{P}}& \textbf{\textit{F1}} \\
					\midrule[1.5pt]
					\textbf{PV-DBOW}  &  & 80.19 & 78.62 & 79.04 & & 83.41 & 81.28 & 82.64 & & 74.36 & 72.91 & 73.19 & & 46.65 & 44.80 & 45.30 \\
					\textbf{fastText}  &  & 86.46 & 85.31 & 84.08 & & 82.98 & 80.12 & 81.73 & & 79.38 & 75.67 & 78.13 & & 57.70 & 53.14 & 56.31 \\
					\textbf{TextCNN}  &  & 92.73 & 90.05 & 92.40 & & 93.64 & 92.72 & 93.25 & & 76.78 & 73.64 & 76.42 & & 43.87 & 41.62 & 43.48 \\
					\textbf{RNN-Capsule} &  & 85.52 & 84.25 & 83.21 & & 86.43 & 85.32 & 85.92 & & 73.18 & 72.49 & 73.02 & & 49.37 & 46.98 & 49.10 \\
					\textbf{Bi-LSTM}  &  & 84.71 & 83.15 & 83.16 & & 87.17 & 86.85 & 85.89 & & 84.25 & 83.15 & 83.04 & & 68.53 & 65.47 & 67.92 \\
					\textbf{Bert-large}  &  & 92.03 & 89.36 & 91.85 & & 97.22 & 95.44 & 96.90 & & 79.23 & 78.47 & 79.02 & & 67.45 & 65.76 & 66.87 \\
					\textbf{TextGCN}  &  & 86.92 & 85.47 & 86.51 & & 88.23 & 87.15 & 86.92 & & 85.69 & 83.67 & 85.15 & & 68.36 & 67.52 & 67.92 \\
					\textbf{RGCN}  &  & 94.74 & \textbf{93.21} & 92.33 & & 93.62 & 91.09 & 92.16 & & 78.72 & 77.06 & 77.45 & & 67.51 & 64.78 & 65.90 \\
					\textbf{HAN}\  &  & 86.17 & 84.67 & 83.08 & & 89.06 & 87.36 & 88.52 & & 79.86 & 78.26 & 77.30 & & 68.20 & 65.14 & 67.51 \\
					\textbf{TGNCL}\  &  & 94.10 & 90.12 & 93.27 & & 96.37 & 94.25 & 95.18 & & 85.92 & 84.86 & 85.13 & & 67.82 & 66.47 & 66.03 \\
					\textbf{ConNHS} &  & \textbf{95.86} & 93.14 & \textbf{94.51} & & \textbf{97.52} & \textbf{96.43} & \textbf{96.93} & & \textbf{87.43} & \textbf{85.46} & \textbf{86.98} & & \textbf{70.65} & \textbf{69.01} & \textbf{69.32} \\
					\bottomrule[1.5pt]
				\end{tabular}
			}
			\label{table2}
		\end{center}
	\end{table*}
	
	In a deeper analysis, we observe that there are also differences in classification capabilities between baseline methods. Firstly, the performance of deep learning-based baselines significantly surpassed that of word embedding models. Notably, Bert-large achieved performance competitive with GNN-based methods. This can be attributed to its pretraining on large-scale corpora and the bidirectional attention mechanism to understand each word in context, thereby possessing a strong semantic understanding capability. Besides, thanks to the fact that graph structures can construct relationships between texts, methods based on graph neural networks (including TextGCN, RGCN, and HAN) have achieved more outstanding classification accuracy than methods based on traditional deep learning. It is worth mentioning that TGNCL, a method based on graph contrastive learning, achieved highly competitive results but did not surpass our proposed ConNHS. This finding suggests that the graph augmentation adopted by TGNCL might, to some extent, disrupt critical semantic information in the text.

	\captionsetup[table]{skip=0pt}
	\begin{table*}[h]
		
		\renewcommand\arraystretch{1.3}
		\setlength{\tabcolsep}{5pt}
		\caption{Test accuracy(\%) and F1 score(\%) for different models with multi-relational text graph.}
		
		\begin{center}
			
			\resizebox{\linewidth}{!}
			%	\scalebox{0.8}
			%\setlength{\tabcolsep}{3.7mm}
			{\fontsize{10pt}{4pt}
				\begin{tabular}{ccccccccccccc}
					\toprule[1.5pt]
					\multirow{2}{*}{\textbf{Method}}&&\multicolumn{2}{c}{\textbf{ThuCNews}}&&\multicolumn{2}{c}{\textbf{SogouNews}}&&\multicolumn{2}{c}{\textbf{20NG}}&&\multicolumn{2}{c}{\textbf{Ohsumed}} \\
					\cline{3-4} \cline{6-7} \cline{9-10} \cline{12-13}
					~ && \textbf{\textit{Acc}}& \textbf{\textit{F1}} && \textbf{\textit{Acc}}& \textbf{\textit{F1}} && \textbf{\textit{Acc}}& \textbf{\textit{F1}} && \textbf{\textit{Acc}}& \textbf{\textit{F1}} \\
					\midrule[1.5pt]
					
					\textbf{RGCN}\  &  & 94.74 & 92.33 & & 93.62 & 92.16 & & 78.72 & 77.45 & & 67.51 & 65.90 \\
					
					\textbf{RGCN\_MTG}\  &  & 94.95(+0.21) & 92.46(+0.13) & & 93.80(+0.18) & 92.21(+0.05) & & 80.97(+2.25) & 79.29(+1.84) & & 69.11(+1.60) & 67.21(+1.31) \\
					
					\textbf{HAN}\  &  & 86.17 & 83.08 & & 89.06 & 88.52 & & 79.86 & 77.30 & & 68.20 & 67.51 \\
					
					\textbf{HAN\_MTG}\  &  & 90.42(+4.25) & 88.61(+5.53) & & 92.18(+3.12) & 91.53(+3.01) & & 81.97(+2.11) & 78.56(+1.26) & & 69.15(+0.95) & 67.71(+0.20) \\
					
					\textbf{ConNHS} &  & \textbf{95.86} & \textbf{94.51} & & \textbf{97.52} & \textbf{96.93} & & \textbf{87.43} & \textbf{86.98} & & \textbf{70.65} & \textbf{69.32} \\
					\bottomrule[1.5pt]
				\end{tabular}
			}
			\label{table3}
		\end{center}
	\end{table*}
	
	\subsubsection{The effectiveness of the Multi-relational Text Graph}
	To assess the effectiveness of our proposed multi-relational text graph (MTG) on text classification, we integrate MTG with other graph neural network models to observe the variations in text classification results.
	
	The experimental results shown in Table \ref{table3} indicate that: by constructing semantic relationships based on core textual features, our multi-relational text graph effectively facilitates nodes learning richer semantic information from their diverse neighbors, thereby generating superior text representations. Upon leveraging our proposed multi-relational text graph, both the HAN and RGCN models exhibit an enhancement in classification accuracy performance. Notably, the HAN model has demonstrated a pronounced improvement in its performance across the two Chinese datasets, registering an accuracy gain exceeding 3\%. The results manifest that despite the HAN model's capacity to adaptively acquire node representations via a dual attention mechanism, yielding excellent performance, the incorporation of multiple semantic relationships among documents offers additional perspectives, thereby enabling the HAN to capture more high-dimensional semantic information, consequently enhancing its learning capabilities. Conversely, the combination of the multi-relational text graph with the RGCN yields more substantial improvements in English datasets. While RGCN already showcases commendable performance on Chinese datasets, the integration of our proposed text graph still facilitates a certain level of advancement. By initiating the process with the extraction of core textual features and establishing multiple inter-text relationships, our approach effectively encourages models to assimilate and interpret high-dimensional semantic information. These experimental findings highlight the intrinsic value of multi-relational text graphs in enhancing text classification tasks.
	
	\subsubsection{Text classification with few labels}
	The advantage of self-supervised GCL lies in its ability to train models using unlabeled data when labels are inaccessible or scarce. Our proposed ConNHS is designed for semi-supervised text classification tasks, requiring ground-true text labels during the testing phase. Therefore, we further test the performance of ConNHS in semi-supervised text classification under conditions of low label availability. We select different proportions of labelled data on the 20NG dataset to assess Bi-LSTM, TextGCN and the proposed ConNHS. To simulate scenarios of scarce labels, we set label rates at 1\%, 2\%, 5\%, and 10\%.	
		
\begin{figure}[!h]
		\centering
		\includegraphics[width=1.0\linewidth,scale=1.00]{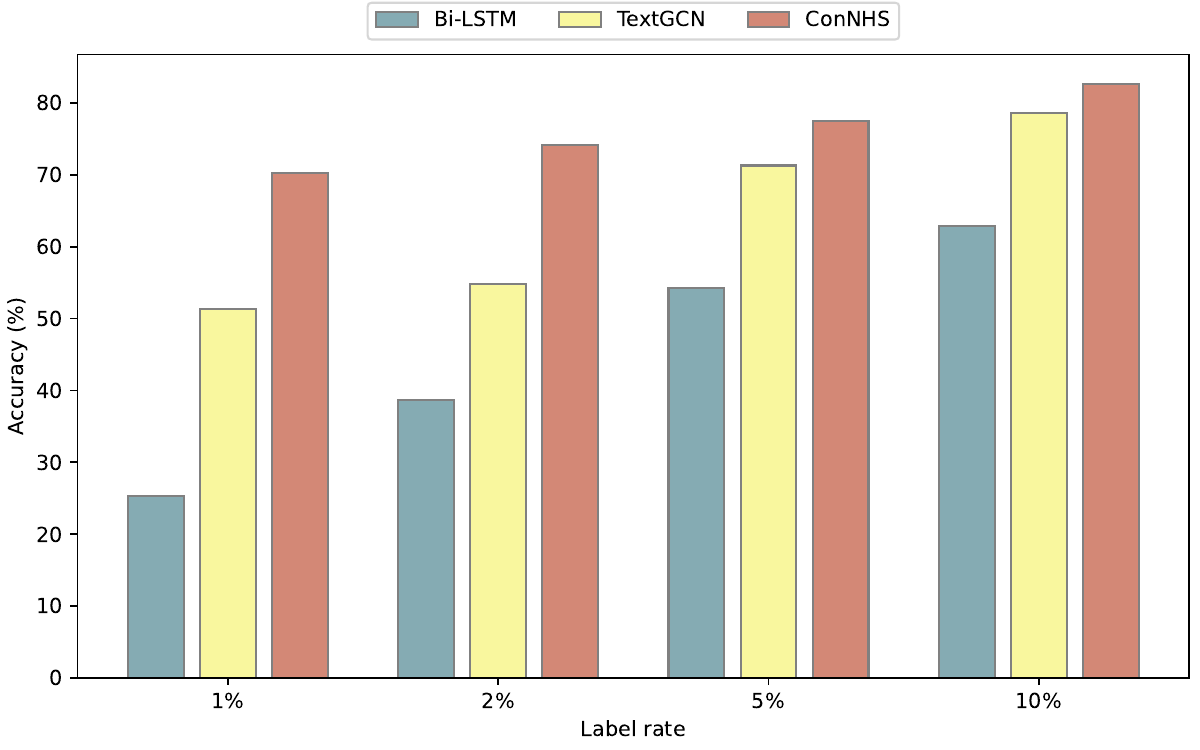}
		\caption{The accuracy with few labels}
		\label{Fig3}
	\end{figure}

	The results in Figure \ref{Fig3} indicate that under conditions of low label rates, our proposed ConNHS exhibits superior classification performance. It is noteworthy that with only a sparse 1\% of labelled text, our method still achieved an accuracy of 70.21\%, while Bi-LSTM and TextGCN experienced a significant drop in performance. The reason behind this is that ConNHS effectively leverages large amounts of unlabeled data for training through self-supervised Graph Contrastive Learning (GCL). In contrast, Bi-LSTM and TextGCN do not incorporate any samples from the test set (unlabeled data) during the computation of training loss. The classification results with few labels indicate that, even with very sparse labeled text, our proposed method can be effectively applied to semi-supervised text classification tasks.

	\subsection{Ablation studies}
	To validate the effectiveness of our proposed contrastive loss NHS, this study conducted a series of ablation experiments on the ThuCnews, SogouNews, 20NG, and Ohsumed datasets. We design different experimental setups for the ablation study: employing the NT-Xent loss, removing the structure-guided signal, removing the attribute-guided signal, and utilizing the complete loss NHS. In these settings, NHS-na represents the removal of node attribute information as the guiding signal for negative sampling, leading to situations where high-order neighbors with high similarity in the graph might still be considered negative samples. NHS-gs denotes disregarding the graph homophily assumption, treating first-order neighbors of the anchor node as negative samples. Furthermore, NT-Xent, a well-established contrastive loss in graph contrastive learning, differs from NHS in that it considers both first-order neighbors and high-order similar neighbors of the anchor as negative samples. Through these ablation experiments, we aim to delve into how each component of the NHS specifically impacts model performance.
	
	\captionsetup[figure]{skip=0pt}
	
	\captionsetup[table]{skip=0pt}
	\begin{table*}[h]
		
		\renewcommand\arraystretch{1.3}
		\setlength{\tabcolsep}{5pt}
		\caption{Ablation experiment of NHS.}
		
		\begin{center}
			
			\resizebox{\linewidth}{!}
			%\scalebox{1.0}{
				%\setlength{\tabcolsep}{3.7mm}
				{\fontsize{6pt}{8pt}\selectfont
					\begin{tabular}{ccccccccccccc}
						\toprule[0.8pt]
						\multirow{2}{*}{\textbf{Method}}&&\multicolumn{2}{c}{\textbf{ThuCNews}}&&\multicolumn{2}{c}{\textbf{SogouNews}}&&\multicolumn{2}{c}{\textbf{20NG}}&&\multicolumn{2}{c}{\textbf{Ohsumed}} \\
						\cline{3-4} \cline{6-7} \cline{9-10} \cline{12-13}
						~ && \textbf{\textit{Acc}}& \textbf{\textit{F1}} && \textbf{\textit{Acc}}& \textbf{\textit{F1}} && \textbf{\textit{Acc}}& \textbf{\textit{F1}} && \textbf{\textit{Acc}}& \textbf{\textit{F1}} \\
						\midrule[0.8pt]
						
						\textbf{NT-Xent}\  &  & 92.31 & 89.02 & & 93.67 & 92.51 & & 84.56 & 83.45 & & 68.32 & 67.18 \\
						
						\textbf{NHS} &  & \textbf{95.86} & \textbf{94.51} & & \textbf{97.52} & \textbf{96.93} & & \textbf{87.43} & \textbf{86.98} & & \textbf{70.65} & \textbf{69.32} \\
						
						\textbf{NHS-gs}\  &  & 94.38 & 93.34 & & 95.56 & 95.08 & & 85.63 & 85.34 & & 69.56 & 68.41 \\
						
						\textbf{NHS-na}\  &  & 95.04 & 94.82 & & 96.21 & 95.69 & & 86.27 & 85.97 & & 69.94 & 68.84 \\
						
						\bottomrule[0.8pt]
					\end{tabular}
				}
				\label{table4}
			\end{center}
		\end{table*}
	
	As shown in the ablation study results in Table \ref{table4}, we observed that employing the NHS contrastive loss achieved the best performance on all four datasets. A decline in classification performance is noted when varying the contrastive loss of the ConNHS method, further highlighting the critical role of our proposed NHS contrastive loss in text classification tasks. Specifically, when switching to the NT-Xent contrastive loss, there is a decline in classification accuracy ranging between 3.33\% to 5.55\%. This result suggests that treating all remaining nodes in the graph as negatives inevitably increases the distance between similar document nodes in the embedding space, thereby reducing the accuracy of text classification. On the other hand, removing the graph structure information as the supervisory signal for negative sampling results in a decrease in accuracy ranging between 1.03\% and 1.96\%. Similarly, removing node attribute information also led to a certain degree of performance decline. Notably, the former scenario caused a more pronounced performance drop than the latter across all datasets. The underlying reason for this phenomenon is that the construction of the multi-relational text graph is based on the assumption of graph homophily, which posits that document nodes connected tend to have more similar core features and are more likely to belong to the same category. Therefore, excluding first-order neighbors of the anchor node from negative samples according to graph structure information aligns more closely with the objectives of graph contrastive learning. Overall, the results of the ablation experiments across different datasets conclusively demonstrate that our proposed NHS contrastive loss effectively mitigates false negative pairs and enhances the accuracy of text classification tasks.

	\subsection{Parameters sensitivity}
	In this section, we focus on exploring how various key parameters influence the performance of our method. It is worth pointing out that, inspired by the adjustment strategys of hyperparameters \citep{mo2022simple, liu2024g, zhao2023textgcl}, we fix other hyperparameters as constants when investigating the impact of a particular hyperparameter on the performance of ConNHS. This allows for a direct observation of the impact of each hyperparameter on the model's performance. The details of the hyperparameters are illustrated in Table \ref{tab:hyperparameters}.
	
	\begin{table}
		\caption{Various hyperparameters.}
		\renewcommand\arraystretch{0.8}
		\setlength\tabcolsep{1mm}%调列距
		\centering
		\resizebox{1\linewidth}{!}{
			\begin{tabular}{@{}c!{\vline}p{0.8\linewidth}@{}}
				%\arrayrulecolor{hc}
				\toprule[1.3pt]
				\textbf{Hyperparameter} &  \textbf{Impact}  \\ \midrule[1.3pt]
				\bm{$\rho_t$}, \bm{$\rho_e$}, \bm{$\rho_k$} & These parameters determine whether there are similarities in titles, events, and keywords within the text. Their possible values range from [0.3, 0.9].
				\\  \midrule
				\bm{$\gamma_e$}, \bm{$\gamma_k$} & They individually dictate the degree of correlation in event relationships and keyword associations within the text. The value range for \bm{$\gamma_e$} is [3, 8], while \bm{$\gamma_k$} ranges from [5, 11]. \\       
				\\ \midrule
				\bm{$\tau$} & It regulates the model's sensitivity to variations in similarity. The adjustment range for \bm{$\tau$} is between 0.05 and 1.0.
				\\ \bottomrule[1.3pt]
		\end{tabular}}
		\label{tab:hyperparameters}
	\end{table}

	\subsubsection{The impact of similarity threshold}
	Selecting an appropriate similarity threshold is vital for constructing a multi-relational text graph, as the establishment of the text graph is highly dependent on the degree of similarity between core features. To investigate the impact of changes in the similarity threshold on the performance of our method, we conducted a series of experiments and visualized the results for detailed analysis and reference. We performed independent experiments by sequentially varying the values of the title threshold, event threshold, and keyword threshold.
	
\begin{figure*}
		\centering
		\includegraphics[width=0.99 \textwidth]{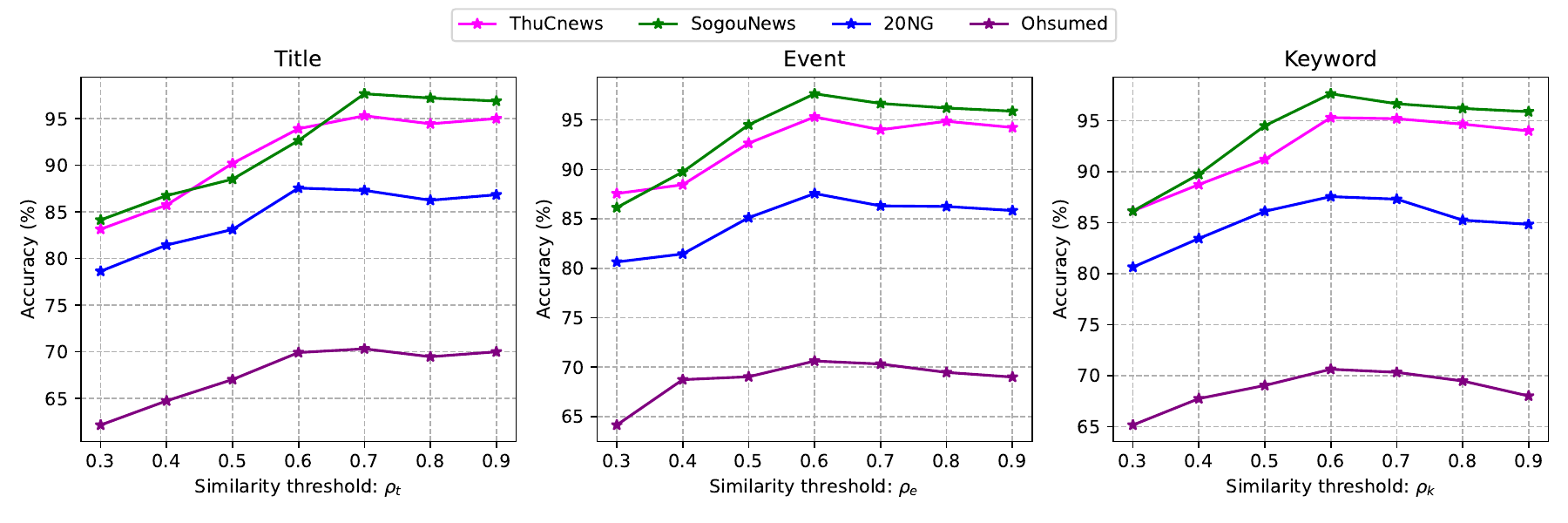}
		\caption{The ConNHS performance under different similarity threshold of core features}
		\label{Fig4}
	\end{figure*}

	As demonstrated in Figure \ref{Fig4}, we observed a clear trend that the classification accuracy tends to increase as the $\rho_t$ rises. However, it is noteworthy that once the $\rho_t$ exceeds 0.7, the improvement in accuracy becomes more gradual. In terms of event feature analysis, increasing the $\rho_e$ indeed effectively boosts accuracy, a trend that continues until the threshold reaches 0.6. Beyond this point, further increases in the $\rho_e$ lead to a decrease in accuracy. This indicates that overly high similarity thresholds might reduce the connections between similar texts, weakening the model's ability to learn textual information under event relations. Additionally, we found that increasing the $\rho_k$ also enhances classification accuracy, but this trend reverses when the $\rho_k$ exceeds 0.6. For all core features, the model performs worse when the similarity threshold is too low. This may be because a low threshold creates redundant edges between text nodes. An appropriate threshold, on the other hand, establishes more reliable connections, thereby improving the node representations learned by the graph neural network. Experimental results indicate that the optimal $\rho_t$ is 0.7. While $\rho_e$ is 0.6, ConNHS achieved the best performance across all four datasets with these settings. For $\rho_k$, the optimal threshold range is between 0.6 and 0.7.
	
	\subsubsection{The impact of minimum association coefficient}
	Texts that share semantically similar events or keywords tend to belong to the same domain. Therefore, we evaluate the impact of different minimum association coefficients $\gamma_e$ and $\gamma_k$ on the performance of the proposed ConNHS in text classification tasks.
	
\begin{figure*}
		\centering
		\includegraphics[scale=0.6]{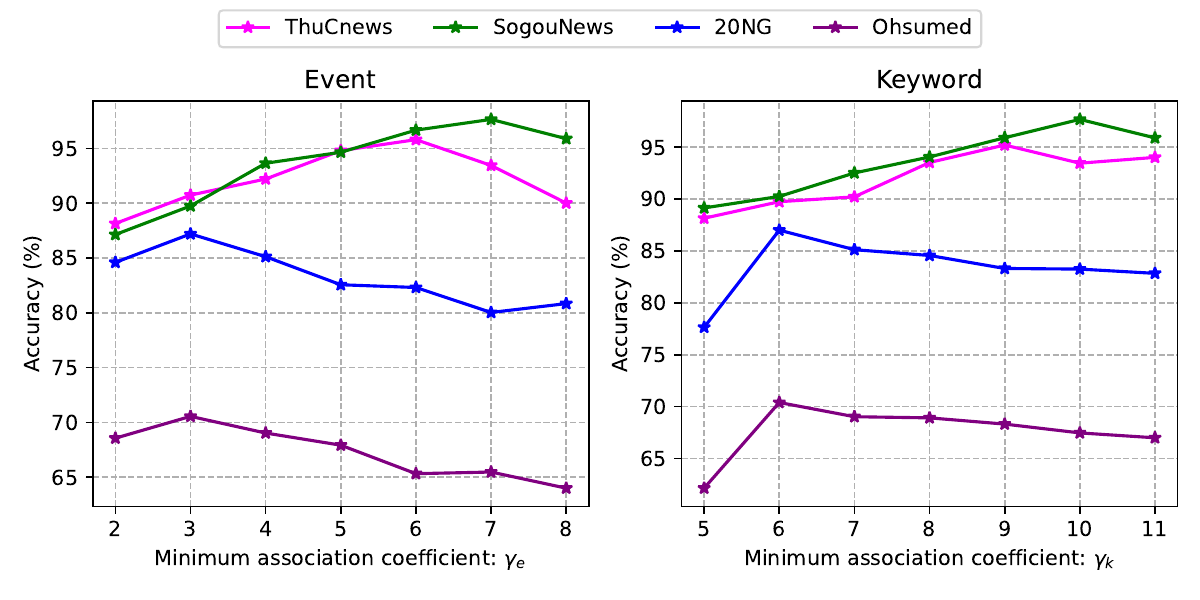}
		\caption{The ConNHS performance under different minimum association coefficient}
		\label{Fig5}
	\end{figure*}

	As shown in Figure \ref{Fig5}, we observe that the accuracy of text classification increases with the rise of $\gamma_e$ and $\gamma_k$. Overall, compared to the Chinese dataset, the two English datasets, which have shorter average lengths, achieve optimal results more quickly. Specifically, when the value of $\gamma_e$ is 3, ConNHS achieves the highest classification accuracy on the 20NG and Ohsumed datasets. In contrast, when $\gamma_e$ is set to 6 and 7, the corresponding accuracies for ThuCnews and SogouNews are better. For the minimum association threshold $\gamma_k$, when its value is 6, the 20NG and Ohsumed datasets achieve the best results. In contrast, ThuCnews and SogouNews achieve the highest classification accuracy when $\gamma_k$ is set to 9 and 10, respectively. It is worth noting that for shorter datasets, after reaching the highest accuracy, further increasing the values of $\gamma_e$ and $\gamma_k$ leads to a significant decline in performance. The experimental results reveal a trend that for longer datasets, the optimal values of $\gamma_e$ and $\gamma_k$ tend to be higher than those for shorter datasets. This is because short texts have limited feature information, and setting the minimum association threshold too high may cause many potential semantic connections to be overlooked, thereby reducing effective links between texts.

	\subsubsection{The impact of temperature hyperparameter}
\begin{figure*}
		\centering
		\includegraphics[width=0.65 \textwidth]{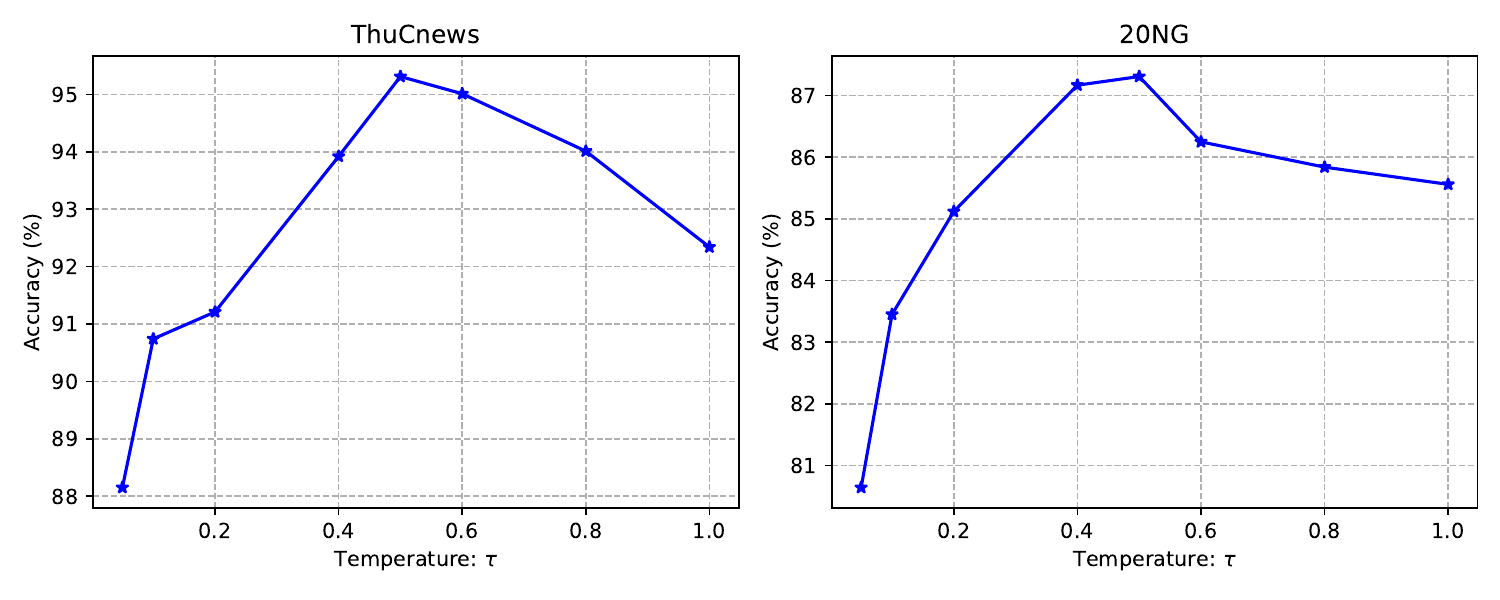}
		\caption{The ConNHS performance under different temperature}
		\label{Fig6}
	\end{figure*}

	The temperature parameter plays a pivotal role in graph contrastive learning as a fundamental hyperparameter that modulates the distribution of similarity scores within the contrastive loss function. To analyze the impact of the temperature parameter on classification accuracy, we conduct validation on the ThuCnews and 20NG datasets. As shown in Figure \ref{Fig6}, the results indicate that a too-low temperature parameter leads to suboptimal classification outcomes. As the value of $\tau$ increases, the model classification capability improves, and our method achieves the best results on both ThuCnews and 20NG when $\tau$ is approximately 0.5. It is noteworthy that a value that is too high for the temperature parameter can also lead to a decline in performance. The experimental results suggest that the value of $\tau$ may require fine-tuning for different datasets to achieve optimal performance. In general, we recommend starting with a value of 0.5 and conducting a thorough parameter search within the range of 0.4 to 0.7.

	\section{Conclusion}
	\label{sec:conclusion}
	In this paper, the ConNHS method we propose demonstrates competitive performance in semi-supervised text classification tasks. Firstly, inspired by the logic humans use to categorize texts, we constructed a multi-relational text graph. Subsequently, we introduced RW-GCN and CGAN for intra-graph and inter-graph propagation, respectively. RW-GCN leverages edge features to capture varying correlations between nodes, while CGAN learns the differences in inter-graph features and integrates document node representations. Additionally, we introduced the neighbor hierarchical sifting  loss to optimize the negative selection process, effectively mitigating the issue of false negatives. Extensive experiments conducted on multiple datasets demonstrate that our method achieves superior results across various evaluation metrics compared to existing approaches. It is worth noting that the multi-relational graphs we constructed inevitably contain some noisy edges, which may mislead the learning process of the graph neural networks. In future work, we will explore denoising techniques in multi-relational text graphs to further optimize the node aggregation process and enhance model performance. Meanwhile, we will continue to investigate graph contrastive learning, with a particular focus on optimizing the negative sample selection process.
		
	\section*{CRediT authorship contribution statement}
	\textbf{Wei Ai}: Supervision, Investigation, Writing - Review \& Editing. \textbf{Jianbin Li}: Conceptualization, Methodology, Investigation, Data curation, Writing - Original Draft. \textbf{Ze Wang}: Supervision, Writing - Review \& Editing. \textbf{Yingying Wei}: Supervision, Investigation \& Review. \textbf{Tao Meng}: Supervision, Investigation, Writing - Review \& Editing. \textbf{Keqin Li}: Supervision, Investigation, Writing - Review \& Editing.
	
	\section*{Declaration of Competing Interest}
	The authors declare that they have no known competing financial interests or personal relationships that could have appeared to influence the work reported in this paper.
	
	\section*{Data availability}
	Data will be made available on request.
	
	\section*{Acknowledgements}
	This work is supported by National Natural Science Foundation of China (Grant No. 69189338), and Excellent Young Scholars of Hunan Province of China (Grant No. 22B0275).
		
	
	\bibliographystyle{cas-model2-names}
	
	\bibliography{cas-refs.bib}

\begin{thebibliography}{72}
\expandafter\ifx\csname natexlab\endcsname\relax\def\natexlab#1{#1}\fi
\providecommand{\url}[1]{\texttt{#1}}
\providecommand{\href}[2]{#2}
\providecommand{\path}[1]{#1}
\providecommand{\DOIprefix}{doi:}
\providecommand{\ArXivprefix}{arXiv:}
\providecommand{\URLprefix}{URL: }
\providecommand{\Pubmedprefix}{pmid:}
\providecommand{\doi}[1]{\href{http://dx.doi.org/#1}{\path{#1}}}
\providecommand{\Pubmed}[1]{\href{pmid:#1}{\path{#1}}}
\providecommand{\bibinfo}[2]{#2}
\ifx\xfnm\relax \def\xfnm[#1]{\unskip,\space#1}\fi
%Type = Article
\bibitem[{Ai et~al.(2024a)Ai, Deng, Chen, Du, Meng and Shou}]{ai2024mcsff}
\bibinfo{author}{Ai, W.}, \bibinfo{author}{Deng, W.}, \bibinfo{author}{Chen,
  H.}, \bibinfo{author}{Du, J.}, \bibinfo{author}{Meng, T.},
  \bibinfo{author}{Shou, Y.}, \bibinfo{year}{2024}a.
\newblock \bibinfo{title}{Mcsff: Multi-modal consistency and specificity fusion
  framework for entity alignment}.
\newblock \bibinfo{journal}{arXiv preprint arXiv:2410.14584} .
%Type = Article
\bibitem[{Ai et~al.(2024b)Ai, Li, Wang, Du, Meng, Shou and Li}]{ai2024graph}
\bibinfo{author}{Ai, W.}, \bibinfo{author}{Li, J.}, \bibinfo{author}{Wang, Z.},
  \bibinfo{author}{Du, J.}, \bibinfo{author}{Meng, T.}, \bibinfo{author}{Shou,
  Y.}, \bibinfo{author}{Li, K.}, \bibinfo{year}{2024}b.
\newblock \bibinfo{title}{Graph contrastive learning via cluster-refined
  negative sampling for semi-supervised text classification}.
\newblock \bibinfo{journal}{arXiv preprint arXiv:2410.18130} .
%Type = Article
\bibitem[{Ai et~al.(2024c)Ai, Shou, Meng and Li}]{ai2024gcn}
\bibinfo{author}{Ai, W.}, \bibinfo{author}{Shou, Y.}, \bibinfo{author}{Meng,
  T.}, \bibinfo{author}{Li, K.}, \bibinfo{year}{2024}c.
\newblock \bibinfo{title}{Der-gcn: Dialog and event relation-aware graph
  convolutional neural network for multimodal dialog emotion recognition}.
\newblock \bibinfo{journal}{IEEE Transactions on Neural Networks and Learning
  Systems} .
%Type = Article
\bibitem[{Ai et~al.(2023a)Ai, Shou, Meng, Yin and Li}]{ai2023gcn}
\bibinfo{author}{Ai, W.}, \bibinfo{author}{Shou, Y.}, \bibinfo{author}{Meng,
  T.}, \bibinfo{author}{Yin, N.}, \bibinfo{author}{Li, K.},
  \bibinfo{year}{2023}a.
\newblock \bibinfo{title}{Der-gcn: Dialogue and event relation-aware graph
  convolutional neural network for multimodal dialogue emotion recognition}.
\newblock \bibinfo{journal}{arXiv preprint arXiv:2312.10579} .
%Type = Article
\bibitem[{Ai et~al.(2024d)Ai, Wei, Shao, Shou, Meng and Li}]{ai2024edge}
\bibinfo{author}{Ai, W.}, \bibinfo{author}{Wei, Y.}, \bibinfo{author}{Shao,
  H.}, \bibinfo{author}{Shou, Y.}, \bibinfo{author}{Meng, T.},
  \bibinfo{author}{Li, K.}, \bibinfo{year}{2024}d.
\newblock \bibinfo{title}{Edge-enhanced minimum-margin graph attention network
  for short text classification}.
\newblock \bibinfo{journal}{Expert Systems with Applications}
  \bibinfo{volume}{251}, \bibinfo{pages}{124069}.
%Type = Inproceedings
\bibitem[{Ai et~al.(2023b)Ai, Zhang, Meng, Shou, Shao and Li}]{ai2023two}
\bibinfo{author}{Ai, W.}, \bibinfo{author}{Zhang, F.}, \bibinfo{author}{Meng,
  T.}, \bibinfo{author}{Shou, Y.}, \bibinfo{author}{Shao, H.},
  \bibinfo{author}{Li, K.}, \bibinfo{year}{2023}b.
\newblock \bibinfo{title}{A two-stage multimodal emotion recognition model
  based on graph contrastive learning}, in: \bibinfo{booktitle}{2023 IEEE 29th
  International Conference on Parallel and Distributed Systems (ICPADS)},
  \bibinfo{organization}{IEEE}. pp. \bibinfo{pages}{397--404}.
%Type = Inproceedings
\bibitem[{Chang et~al.(2020)Chang, Yu, Zhong, Yang and
  Dhillon}]{chang2020taming}
\bibinfo{author}{Chang, W.C.}, \bibinfo{author}{Yu, H.F.},
  \bibinfo{author}{Zhong, K.}, \bibinfo{author}{Yang, Y.},
  \bibinfo{author}{Dhillon, I.S.}, \bibinfo{year}{2020}.
\newblock \bibinfo{title}{Taming pretrained transformers for extreme
  multi-label text classification}, in: \bibinfo{booktitle}{Proceedings of the
  26th ACM SIGKDD international conference on knowledge discovery \& data
  mining}, pp. \bibinfo{pages}{3163--3171}.
%Type = Article
\bibitem[{Chen et~al.(2024)Chen, Xiao, Zhang, Luo, Lian and Liu}]{chen2024bge}
\bibinfo{author}{Chen, J.}, \bibinfo{author}{Xiao, S.}, \bibinfo{author}{Zhang,
  P.}, \bibinfo{author}{Luo, K.}, \bibinfo{author}{Lian, D.},
  \bibinfo{author}{Liu, Z.}, \bibinfo{year}{2024}.
\newblock \bibinfo{title}{Bge m3-embedding: Multi-lingual, multi-functionality,
  multi-granularity text embeddings through self-knowledge distillation}.
\newblock \bibinfo{journal}{arXiv preprint arXiv:2402.03216} .
%Type = Article
\bibitem[{Dieng et~al.(2016)Dieng, Wang, Gao and Paisley}]{dieng2016topicrnn}
\bibinfo{author}{Dieng, A.B.}, \bibinfo{author}{Wang, C.},
  \bibinfo{author}{Gao, J.}, \bibinfo{author}{Paisley, J.},
  \bibinfo{year}{2016}.
\newblock \bibinfo{title}{Topicrnn: A recurrent neural network with long-range
  semantic dependency}.
\newblock \bibinfo{journal}{arXiv preprint arXiv:1611.01702} .
%Type = Inproceedings
\bibitem[{Hassani and Khasahmadi(2020)}]{hassani2020contrastive}
\bibinfo{author}{Hassani, K.}, \bibinfo{author}{Khasahmadi, A.H.},
  \bibinfo{year}{2020}.
\newblock \bibinfo{title}{Contrastive multi-view representation learning on
  graphs}, in: \bibinfo{booktitle}{International conference on machine
  learning}, \bibinfo{organization}{PMLR}. pp. \bibinfo{pages}{4116--4126}.
%Type = Inproceedings
\bibitem[{Huang et~al.(2019)Huang, Ma, Li, Zhang and Wang}]{huang2019text}
\bibinfo{author}{Huang, L.}, \bibinfo{author}{Ma, D.}, \bibinfo{author}{Li,
  S.}, \bibinfo{author}{Zhang, X.}, \bibinfo{author}{Wang, H.},
  \bibinfo{year}{2019}.
\newblock \bibinfo{title}{Text level graph neural network for text
  classification}, in: \bibinfo{booktitle}{Proceedings of the 2019 Conference
  on Empirical Methods in Natural Language Processing and the 9th International
  Joint Conference on Natural Language Processing (EMNLP-IJCNLP)}, pp.
  \bibinfo{pages}{3444--3450}.
%Type = Inproceedings
\bibitem[{Joulin et~al.(2017)Joulin, Grave, Bojanowski and
  Mikolov}]{joulin2017bag}
\bibinfo{author}{Joulin, A.}, \bibinfo{author}{Grave, {\'E}.},
  \bibinfo{author}{Bojanowski, P.}, \bibinfo{author}{Mikolov, T.},
  \bibinfo{year}{2017}.
\newblock \bibinfo{title}{Bag of tricks for efficient text classification}, in:
  \bibinfo{booktitle}{Proceedings of the 15th Conference of the European
  Chapter of the Association for Computational Linguistics: Volume 2, Short
  Papers}, pp. \bibinfo{pages}{427--431}.
%Type = Inproceedings
\bibitem[{Kenton and Toutanova(2019)}]{kenton2019bert}
\bibinfo{author}{Kenton, J.D.M.W.C.}, \bibinfo{author}{Toutanova, L.K.},
  \bibinfo{year}{2019}.
\newblock \bibinfo{title}{Bert: Pre-training of deep bidirectional transformers
  for language understanding}, in: \bibinfo{booktitle}{Proceedings of
  NAACL-HLT}, pp. \bibinfo{pages}{4171--4186}.
%Type = Article
\bibitem[{Kim(2014)}]{kim2014convolutional}
\bibinfo{author}{Kim, Y.}, \bibinfo{year}{2014}.
\newblock \bibinfo{title}{Convolutional neural networks for sentence
  classification}.
\newblock \bibinfo{journal}{arXiv preprint arXiv:1408.5882} .
%Type = Inproceedings
\bibitem[{Kipf and Welling(2016)}]{kipf2016semi}
\bibinfo{author}{Kipf, T.N.}, \bibinfo{author}{Welling, M.},
  \bibinfo{year}{2016}.
\newblock \bibinfo{title}{Semi-supervised classification with graph
  convolutional networks}, in: \bibinfo{booktitle}{International Conference on
  Learning Representations}, pp. \bibinfo{pages}{1--10}.
%Type = Inproceedings
\bibitem[{Lai et~al.(2015)Lai, Xu, Liu and Zhao}]{lai2015recurrent}
\bibinfo{author}{Lai, S.}, \bibinfo{author}{Xu, L.}, \bibinfo{author}{Liu, K.},
  \bibinfo{author}{Zhao, J.}, \bibinfo{year}{2015}.
\newblock \bibinfo{title}{Recurrent convolutional neural networks for text
  classification}, in: \bibinfo{booktitle}{Proceedings of the AAAI conference
  on artificial intelligence}, pp. \bibinfo{pages}{1--10}.
%Type = Article
\bibitem[{Lan et~al.(2023)Lan, Hu, Li and Zhang}]{lan2023contrastive}
\bibinfo{author}{Lan, G.}, \bibinfo{author}{Hu, M.}, \bibinfo{author}{Li, Y.},
  \bibinfo{author}{Zhang, Y.}, \bibinfo{year}{2023}.
\newblock \bibinfo{title}{Contrastive knowledge integrated graph neural
  networks for chinese medical text classification}.
\newblock \bibinfo{journal}{Engineering Applications of Artificial
  Intelligence} \bibinfo{volume}{122}, \bibinfo{pages}{106057}.
%Type = Inproceedings
\bibitem[{Le and Mikolov(2014)}]{le2014distributed}
\bibinfo{author}{Le, Q.}, \bibinfo{author}{Mikolov, T.}, \bibinfo{year}{2014}.
\newblock \bibinfo{title}{Distributed representations of sentences and
  documents}, in: \bibinfo{booktitle}{International conference on machine
  learning}, \bibinfo{organization}{PMLR}. pp. \bibinfo{pages}{1188--1196}.
%Type = Article
\bibitem[{Lei et~al.(2021)Lei, Liu, Li, Dai and Wang}]{lei2021multihop}
\bibinfo{author}{Lei, F.}, \bibinfo{author}{Liu, X.}, \bibinfo{author}{Li, Z.},
  \bibinfo{author}{Dai, Q.}, \bibinfo{author}{Wang, S.}, \bibinfo{year}{2021}.
\newblock \bibinfo{title}{Multihop neighbor information fusion graph
  convolutional network for text classification}.
\newblock \bibinfo{journal}{Mathematical Problems in Engineering}
  \bibinfo{volume}{2021}, \bibinfo{pages}{1--9}.
%Type = Inproceedings
\bibitem[{Li et~al.(2021)Li, Peng, Peng, Li and Wang}]{li2021textgtl}
\bibinfo{author}{Li, C.}, \bibinfo{author}{Peng, X.}, \bibinfo{author}{Peng,
  H.}, \bibinfo{author}{Li, J.}, \bibinfo{author}{Wang, L.},
  \bibinfo{year}{2021}.
\newblock \bibinfo{title}{Textgtl: Graph-based transductive learning for
  semi-supervised text classification via structure-sensitive interpolation.},
  in: \bibinfo{booktitle}{IJCAI}, pp. \bibinfo{pages}{2680--2686}.
%Type = Article
\bibitem[{Li et~al.(2023)Li, Wang, Wang and Wang}]{li2023graph}
\bibinfo{author}{Li, X.}, \bibinfo{author}{Wang, B.}, \bibinfo{author}{Wang,
  Y.}, \bibinfo{author}{Wang, M.}, \bibinfo{year}{2023}.
\newblock \bibinfo{title}{Graph-based text classification by contrastive
  learning with text-level graph augmentation}.
\newblock \bibinfo{journal}{ACM Transactions on Knowledge Discovery from Data}
  .
%Type = Inproceedings
\bibitem[{Lin et~al.(2021)Lin, Meng, Sun, Han, Kuang, Li and
  Wu}]{lin2021bertgcn}
\bibinfo{author}{Lin, Y.}, \bibinfo{author}{Meng, Y.}, \bibinfo{author}{Sun,
  X.}, \bibinfo{author}{Han, Q.}, \bibinfo{author}{Kuang, K.},
  \bibinfo{author}{Li, J.}, \bibinfo{author}{Wu, F.}, \bibinfo{year}{2021}.
\newblock \bibinfo{title}{Bertgcn: Transductive text classification by
  combining gnn and bert}, in: \bibinfo{booktitle}{Findings of the Association
  for Computational Linguistics: ACL-IJCNLP 2021}, pp.
  \bibinfo{pages}{1456--1462}.
%Type = Inproceedings
\bibitem[{Linmei et~al.(2019)Linmei, Yang, Shi, Ji and
  Li}]{linmei2019heterogeneous}
\bibinfo{author}{Linmei, H.}, \bibinfo{author}{Yang, T.}, \bibinfo{author}{Shi,
  C.}, \bibinfo{author}{Ji, H.}, \bibinfo{author}{Li, X.},
  \bibinfo{year}{2019}.
\newblock \bibinfo{title}{Heterogeneous graph attention networks for
  semi-supervised short text classification}, in:
  \bibinfo{booktitle}{Proceedings of the 2019 conference on empirical methods
  in natural language processing and the 9th international joint conference on
  natural language processing (EMNLP-IJCNLP)}, pp. \bibinfo{pages}{4821--4830}.
%Type = Article
\bibitem[{Liu et~al.(2024)Liu, Ma, Wei, Ji, Yang and Abraham}]{liu2024g}
\bibinfo{author}{Liu, X.}, \bibinfo{author}{Ma, K.}, \bibinfo{author}{Wei, Q.},
  \bibinfo{author}{Ji, K.}, \bibinfo{author}{Yang, B.},
  \bibinfo{author}{Abraham, A.}, \bibinfo{year}{2024}.
\newblock \bibinfo{title}{G-hfin: graph-based hierarchical feature integration
  network for propaganda detection of we-media news articles}.
\newblock \bibinfo{journal}{Engineering Applications of Artificial
  Intelligence} \bibinfo{volume}{132}, \bibinfo{pages}{107922}.
%Type = Inproceedings
\bibitem[{Liu et~al.(2020)Liu, You, Zhang, Wu and Lv}]{liu2020tensor}
\bibinfo{author}{Liu, X.}, \bibinfo{author}{You, X.}, \bibinfo{author}{Zhang,
  X.}, \bibinfo{author}{Wu, J.}, \bibinfo{author}{Lv, P.},
  \bibinfo{year}{2020}.
\newblock \bibinfo{title}{Tensor graph convolutional networks for text
  classification}, in: \bibinfo{booktitle}{Proceedings of the AAAI conference
  on artificial intelligence}, pp. \bibinfo{pages}{8409--8416}.
%Type = Article
\bibitem[{Liu et~al.(2019)Liu, Ott, Goyal, Du, Joshi, Chen, Levy, Lewis,
  Zettlemoyer and Stoyanov}]{liu2019roberta}
\bibinfo{author}{Liu, Y.}, \bibinfo{author}{Ott, M.}, \bibinfo{author}{Goyal,
  N.}, \bibinfo{author}{Du, J.}, \bibinfo{author}{Joshi, M.},
  \bibinfo{author}{Chen, D.}, \bibinfo{author}{Levy, O.},
  \bibinfo{author}{Lewis, M.}, \bibinfo{author}{Zettlemoyer, L.},
  \bibinfo{author}{Stoyanov, V.}, \bibinfo{year}{2019}.
\newblock \bibinfo{title}{Roberta: A robustly optimized bert pretraining
  approach}.
\newblock \bibinfo{journal}{arXiv preprint arXiv:1907.11692} .
%Type = Article
\bibitem[{Meng et~al.(2024a)Meng, Shou, Ai, Du, Liu and Li}]{meng2024multi}
\bibinfo{author}{Meng, T.}, \bibinfo{author}{Shou, Y.}, \bibinfo{author}{Ai,
  W.}, \bibinfo{author}{Du, J.}, \bibinfo{author}{Liu, H.},
  \bibinfo{author}{Li, K.}, \bibinfo{year}{2024}a.
\newblock \bibinfo{title}{A multi-message passing framework based on
  heterogeneous graphs in conversational emotion recognition}.
\newblock \bibinfo{journal}{Neurocomputing} \bibinfo{volume}{569},
  \bibinfo{pages}{127109}.
%Type = Article
\bibitem[{Meng et~al.(2024b)Meng, Shou, Ai, Yin and Li}]{meng2024deep}
\bibinfo{author}{Meng, T.}, \bibinfo{author}{Shou, Y.}, \bibinfo{author}{Ai,
  W.}, \bibinfo{author}{Yin, N.}, \bibinfo{author}{Li, K.},
  \bibinfo{year}{2024}b.
\newblock \bibinfo{title}{Deep imbalanced learning for multimodal emotion
  recognition in conversations}.
\newblock \bibinfo{journal}{IEEE Transactions on Artificial Intelligence} .
%Type = Article
\bibitem[{Meng et~al.(2024c)Meng, Zhang, Shou, Ai, Yin and
  Li}]{meng2024revisiting}
\bibinfo{author}{Meng, T.}, \bibinfo{author}{Zhang, F.}, \bibinfo{author}{Shou,
  Y.}, \bibinfo{author}{Ai, W.}, \bibinfo{author}{Yin, N.},
  \bibinfo{author}{Li, K.}, \bibinfo{year}{2024}c.
\newblock \bibinfo{title}{Revisiting multimodal emotion recognition in
  conversation from the perspective of graph spectrum}.
\newblock \bibinfo{journal}{arXiv preprint arXiv:2404.17862} .
%Type = Article
\bibitem[{Meng et~al.(2024d)Meng, Zhang, Shou, Shao, Ai and
  Li}]{meng2024masked}
\bibinfo{author}{Meng, T.}, \bibinfo{author}{Zhang, F.}, \bibinfo{author}{Shou,
  Y.}, \bibinfo{author}{Shao, H.}, \bibinfo{author}{Ai, W.},
  \bibinfo{author}{Li, K.}, \bibinfo{year}{2024}d.
\newblock \bibinfo{title}{Masked graph learning with recurrent alignment for
  multimodal emotion recognition in conversation}.
\newblock \bibinfo{journal}{IEEE/ACM Transactions on Audio, Speech, and
  Language Processing} .
%Type = Article
\bibitem[{Miao et~al.(2022)Miao, Yang, Ma, Juan, Xue, Tang, Wang and
  Wang}]{miao2022negative}
\bibinfo{author}{Miao, R.}, \bibinfo{author}{Yang, Y.}, \bibinfo{author}{Ma,
  Y.}, \bibinfo{author}{Juan, X.}, \bibinfo{author}{Xue, H.},
  \bibinfo{author}{Tang, J.}, \bibinfo{author}{Wang, Y.},
  \bibinfo{author}{Wang, X.}, \bibinfo{year}{2022}.
\newblock \bibinfo{title}{Negative samples selecting strategy for graph
  contrastive learning}.
\newblock \bibinfo{journal}{Information Sciences} \bibinfo{volume}{613},
  \bibinfo{pages}{667--681}.
%Type = Inproceedings
\bibitem[{Mo et~al.(2022)Mo, Peng, Xu, Shi and Zhu}]{mo2022simple}
\bibinfo{author}{Mo, Y.}, \bibinfo{author}{Peng, L.}, \bibinfo{author}{Xu, J.},
  \bibinfo{author}{Shi, X.}, \bibinfo{author}{Zhu, X.}, \bibinfo{year}{2022}.
\newblock \bibinfo{title}{Simple unsupervised graph representation learning},
  in: \bibinfo{booktitle}{Proceedings of the AAAI conference on artificial
  intelligence}, pp. \bibinfo{pages}{7797--7805}.
%Type = Inproceedings
\bibitem[{Piao et~al.(2022)Piao, Lee, Lee and Kim}]{piao2022sparse}
\bibinfo{author}{Piao, Y.}, \bibinfo{author}{Lee, S.}, \bibinfo{author}{Lee,
  D.}, \bibinfo{author}{Kim, S.}, \bibinfo{year}{2022}.
\newblock \bibinfo{title}{Sparse structure learning via graph neural networks
  for inductive document classification}, in: \bibinfo{booktitle}{Proceedings
  of the AAAI Conference on Artificial Intelligence}, pp.
  \bibinfo{pages}{11165--11173}.
%Type = Inproceedings
\bibitem[{Schlichtkrull et~al.(2018)Schlichtkrull, Kipf, Bloem, Van Den~Berg,
  Titov and Welling}]{schlichtkrull2018modeling}
\bibinfo{author}{Schlichtkrull, M.}, \bibinfo{author}{Kipf, T.N.},
  \bibinfo{author}{Bloem, P.}, \bibinfo{author}{Van Den~Berg, R.},
  \bibinfo{author}{Titov, I.}, \bibinfo{author}{Welling, M.},
  \bibinfo{year}{2018}.
\newblock \bibinfo{title}{Modeling relational data with graph convolutional
  networks}, in: \bibinfo{booktitle}{The Semantic Web: 15th International
  Conference, ESWC 2018, Heraklion, Crete, Greece, June 3--7, 2018, Proceedings
  15}, \bibinfo{organization}{Springer}. pp. \bibinfo{pages}{593--607}.
%Type = Inproceedings
\bibitem[{Shen et~al.(2023)Shen, Sun, Pan, Zhou and Yang}]{shen2023neighbor}
\bibinfo{author}{Shen, X.}, \bibinfo{author}{Sun, D.}, \bibinfo{author}{Pan,
  S.}, \bibinfo{author}{Zhou, X.}, \bibinfo{author}{Yang, L.T.},
  \bibinfo{year}{2023}.
\newblock \bibinfo{title}{Neighbor contrastive learning on learnable graph
  augmentation}, in: \bibinfo{booktitle}{Proceedings of the AAAI Conference on
  Artificial Intelligence}, pp. \bibinfo{pages}{9782--9791}.
%Type = Article
\bibitem[{Shi et~al.(2024)Shi, Hu, Xie, Guo and Wu}]{shi2024robust}
\bibinfo{author}{Shi, S.}, \bibinfo{author}{Hu, K.}, \bibinfo{author}{Xie, J.},
  \bibinfo{author}{Guo, Y.}, \bibinfo{author}{Wu, H.}, \bibinfo{year}{2024}.
\newblock \bibinfo{title}{Robust scientific text classification using prompt
  tuning based on data augmentation with l2 regularization}.
\newblock \bibinfo{journal}{Information Processing \& Management}
  \bibinfo{volume}{61}, \bibinfo{pages}{103531}.
%Type = Article
\bibitem[{Shou et~al.(2024a)Shou, Ai, Du, Meng and Liu}]{shou2024efficient}
\bibinfo{author}{Shou, Y.}, \bibinfo{author}{Ai, W.}, \bibinfo{author}{Du, J.},
  \bibinfo{author}{Meng, T.}, \bibinfo{author}{Liu, H.}, \bibinfo{year}{2024}a.
\newblock \bibinfo{title}{Efficient long-distance latent relation-aware graph
  neural network for multi-modal emotion recognition in conversations}.
\newblock \bibinfo{journal}{arXiv preprint arXiv:2407.00119} .
%Type = Article
\bibitem[{Shou et~al.(2023a)Shou, Ai, Meng and Li}]{shou2023czl}
\bibinfo{author}{Shou, Y.}, \bibinfo{author}{Ai, W.}, \bibinfo{author}{Meng,
  T.}, \bibinfo{author}{Li, K.}, \bibinfo{year}{2023}a.
\newblock \bibinfo{title}{Czl-ciae: Clip-driven zero-shot learning for
  correcting inverse age estimation}.
\newblock \bibinfo{journal}{arXiv preprint arXiv:2312.01758} .
%Type = Article
\bibitem[{Shou et~al.(2023b)Shou, Ai, Meng and Yin}]{shou2023graph}
\bibinfo{author}{Shou, Y.}, \bibinfo{author}{Ai, W.}, \bibinfo{author}{Meng,
  T.}, \bibinfo{author}{Yin, N.}, \bibinfo{year}{2023}b.
\newblock \bibinfo{title}{Graph information bottleneck for remote sensing
  segmentation}.
\newblock \bibinfo{journal}{arXiv preprint arXiv:2312.02545} .
%Type = Inproceedings
\bibitem[{Shou et~al.(2023c)Shou, Ai, Meng, Zhang and Li}]{shou2023graphunet}
\bibinfo{author}{Shou, Y.}, \bibinfo{author}{Ai, W.}, \bibinfo{author}{Meng,
  T.}, \bibinfo{author}{Zhang, F.}, \bibinfo{author}{Li, K.},
  \bibinfo{year}{2023}c.
\newblock \bibinfo{title}{Graphunet: Graph make strong encoders for remote
  sensing segmentation}, in: \bibinfo{booktitle}{2023 IEEE 29th International
  Conference on Parallel and Distributed Systems (ICPADS)},
  \bibinfo{organization}{IEEE}. pp. \bibinfo{pages}{2734--2737}.
%Type = Article
\bibitem[{Shou et~al.(2025)Shou, Cao, Liu and Meng}]{shou2025masked}
\bibinfo{author}{Shou, Y.}, \bibinfo{author}{Cao, X.}, \bibinfo{author}{Liu,
  H.}, \bibinfo{author}{Meng, D.}, \bibinfo{year}{2025}.
\newblock \bibinfo{title}{Masked contrastive graph representation learning for
  age estimation}.
\newblock \bibinfo{journal}{Pattern Recognition} \bibinfo{volume}{158},
  \bibinfo{pages}{110974}.
%Type = Article
\bibitem[{Shou et~al.(2024b)Shou, Cao and Meng}]{shou2024spegcl}
\bibinfo{author}{Shou, Y.}, \bibinfo{author}{Cao, X.}, \bibinfo{author}{Meng,
  D.}, \bibinfo{year}{2024}b.
\newblock \bibinfo{title}{Spegcl: Self-supervised graph spectrum contrastive
  learning without positive samples}.
\newblock \bibinfo{journal}{arXiv preprint arXiv:2410.10365} .
%Type = Article
\bibitem[{Shou et~al.(2024c)Shou, Lan and Cao}]{shou2024contrastive}
\bibinfo{author}{Shou, Y.}, \bibinfo{author}{Lan, H.}, \bibinfo{author}{Cao,
  X.}, \bibinfo{year}{2024}c.
\newblock \bibinfo{title}{Contrastive graph representation learning with
  adversarial cross-view reconstruction and information bottleneck}.
\newblock \bibinfo{journal}{arXiv preprint arXiv:2408.00295} .
%Type = Article
\bibitem[{Shou et~al.(2024d)Shou, Liu, Cao, Meng and Dong}]{shou2024low}
\bibinfo{author}{Shou, Y.}, \bibinfo{author}{Liu, H.}, \bibinfo{author}{Cao,
  X.}, \bibinfo{author}{Meng, D.}, \bibinfo{author}{Dong, B.},
  \bibinfo{year}{2024}d.
\newblock \bibinfo{title}{A low-rank matching attention based cross-modal
  feature fusion method for conversational emotion recognition}.
\newblock \bibinfo{journal}{IEEE Transactions on Affective Computing} .
%Type = Article
\bibitem[{Shou et~al.(2022a)Shou, Meng, Ai, Xie, Liu and Wang}]{shou2022object}
\bibinfo{author}{Shou, Y.}, \bibinfo{author}{Meng, T.}, \bibinfo{author}{Ai,
  W.}, \bibinfo{author}{Xie, C.}, \bibinfo{author}{Liu, H.},
  \bibinfo{author}{Wang, Y.}, \bibinfo{year}{2022}a.
\newblock \bibinfo{title}{Object detection in medical images based on
  hierarchical transformer and mask mechanism}.
\newblock \bibinfo{journal}{Computational Intelligence and Neuroscience}
  \bibinfo{volume}{2022}, \bibinfo{pages}{5863782}.
%Type = Article
\bibitem[{Shou et~al.(2022b)Shou, Meng, Ai, Yang and
  Li}]{shou2022conversational}
\bibinfo{author}{Shou, Y.}, \bibinfo{author}{Meng, T.}, \bibinfo{author}{Ai,
  W.}, \bibinfo{author}{Yang, S.}, \bibinfo{author}{Li, K.},
  \bibinfo{year}{2022}b.
\newblock \bibinfo{title}{Conversational emotion recognition studies based on
  graph convolutional neural networks and a dependent syntactic analysis}.
\newblock \bibinfo{journal}{Neurocomputing} \bibinfo{volume}{501},
  \bibinfo{pages}{629--639}.
%Type = Article
\bibitem[{Shou et~al.(2023d)Shou, Meng, Ai, Yin and Li}]{shou2023adversarial}
\bibinfo{author}{Shou, Y.}, \bibinfo{author}{Meng, T.}, \bibinfo{author}{Ai,
  W.}, \bibinfo{author}{Yin, N.}, \bibinfo{author}{Li, K.},
  \bibinfo{year}{2023}d.
\newblock \bibinfo{title}{Adversarial representation with intra-modal and
  inter-modal graph contrastive learning for multimodal emotion recognition}.
\newblock \bibinfo{journal}{arXiv preprint arXiv:2312.16778} .
%Type = Article
\bibitem[{Shou et~al.(2023e)Shou, Meng, Ai, Yin and Li}]{shou2023comprehensive}
\bibinfo{author}{Shou, Y.}, \bibinfo{author}{Meng, T.}, \bibinfo{author}{Ai,
  W.}, \bibinfo{author}{Yin, N.}, \bibinfo{author}{Li, K.},
  \bibinfo{year}{2023}e.
\newblock \bibinfo{title}{A comprehensive survey on multi-modal conversational
  emotion recognition with deep learning}.
\newblock \bibinfo{journal}{arXiv preprint arXiv:2312.05735} .
%Type = Article
\bibitem[{Shou et~al.(2024e)Shou, Meng, Ai, Zhang, Yin and
  Li}]{shou2024adversarial}
\bibinfo{author}{Shou, Y.}, \bibinfo{author}{Meng, T.}, \bibinfo{author}{Ai,
  W.}, \bibinfo{author}{Zhang, F.}, \bibinfo{author}{Yin, N.},
  \bibinfo{author}{Li, K.}, \bibinfo{year}{2024}e.
\newblock \bibinfo{title}{Adversarial alignment and graph fusion via
  information bottleneck for multimodal emotion recognition in conversations}.
\newblock \bibinfo{journal}{Information Fusion} \bibinfo{volume}{112},
  \bibinfo{pages}{102590}.
%Type = Article
\bibitem[{Shou et~al.(2024f)Shou, Meng, Zhang, Yin and Li}]{shou2024revisiting}
\bibinfo{author}{Shou, Y.}, \bibinfo{author}{Meng, T.}, \bibinfo{author}{Zhang,
  F.}, \bibinfo{author}{Yin, N.}, \bibinfo{author}{Li, K.},
  \bibinfo{year}{2024}f.
\newblock \bibinfo{title}{Revisiting multi-modal emotion learning with broad
  state space models and probability-guidance fusion}.
\newblock \bibinfo{journal}{arXiv preprint arXiv:2404.17858} .
%Type = Article
\bibitem[{Shou et~al.(2024g)Shou, Yan, Yuan, Cao, Zhao and
  Meng}]{shou2024graph}
\bibinfo{author}{Shou, Y.}, \bibinfo{author}{Yan, P.}, \bibinfo{author}{Yuan,
  X.}, \bibinfo{author}{Cao, X.}, \bibinfo{author}{Zhao, Q.},
  \bibinfo{author}{Meng, D.}, \bibinfo{year}{2024}g.
\newblock \bibinfo{title}{Graph domain adaptation with dual-branch encoder and
  two-level alignment for whole slide image-based survival prediction}.
\newblock \bibinfo{journal}{arXiv preprint arXiv:2411.14001} .
%Type = Article
\bibitem[{Sun et~al.(2024)Sun, Cheng, Zhang, Tong and Chai}]{sun2024text}
\bibinfo{author}{Sun, G.}, \bibinfo{author}{Cheng, Y.}, \bibinfo{author}{Zhang,
  Z.}, \bibinfo{author}{Tong, X.}, \bibinfo{author}{Chai, T.},
  \bibinfo{year}{2024}.
\newblock \bibinfo{title}{Text classification with improved word embedding and
  adaptive segmentation}.
\newblock \bibinfo{journal}{Expert Systems with Applications}
  \bibinfo{volume}{238}, \bibinfo{pages}{121852}.
%Type = Inproceedings
\bibitem[{Sun et~al.(2022)Sun, Harit, Cristea, Yu, Shi and
  Al~Moubayed}]{sun2022contrastive}
\bibinfo{author}{Sun, Z.}, \bibinfo{author}{Harit, A.},
  \bibinfo{author}{Cristea, A.I.}, \bibinfo{author}{Yu, J.},
  \bibinfo{author}{Shi, L.}, \bibinfo{author}{Al~Moubayed, N.},
  \bibinfo{year}{2022}.
\newblock \bibinfo{title}{Contrastive learning with heterogeneous graph
  attention networks on short text classification}, in:
  \bibinfo{booktitle}{2022 International Joint Conference on Neural Networks
  (IJCNN)}, \bibinfo{organization}{IEEE}. pp. \bibinfo{pages}{1--6}.
%Type = Inproceedings
\bibitem[{Tai et~al.(2015)Tai, Socher and Manning}]{tai2015improved}
\bibinfo{author}{Tai, K.S.}, \bibinfo{author}{Socher, R.},
  \bibinfo{author}{Manning, C.D.}, \bibinfo{year}{2015}.
\newblock \bibinfo{title}{Improved semantic representations from
  tree-structured long short-term memory networks}, in:
  \bibinfo{booktitle}{Proceedings of the 53rd Annual Meeting of the Association
  for Computational Linguistics and the 7th International Joint Conference on
  Natural Language Processing (Volume 1: Long Papers)}, pp.
  \bibinfo{pages}{1556--1566}.
%Type = Inproceedings
\bibitem[{Veli{\v{c}}kovi{\'c} et~al.(2018)Veli{\v{c}}kovi{\'c}, Cucurull,
  Casanova, Romero, Li{\`o} and Bengio}]{velivckovic2018graph}
\bibinfo{author}{Veli{\v{c}}kovi{\'c}, P.}, \bibinfo{author}{Cucurull, G.},
  \bibinfo{author}{Casanova, A.}, \bibinfo{author}{Romero, A.},
  \bibinfo{author}{Li{\`o}, P.}, \bibinfo{author}{Bengio, Y.},
  \bibinfo{year}{2018}.
\newblock \bibinfo{title}{Graph attention networks}, in:
  \bibinfo{booktitle}{International Conference on Learning Representations},
  pp. \bibinfo{pages}{1--10}.
%Type = Inproceedings
\bibitem[{Wang et~al.(2019)Wang, Ji, Shi, Wang, Ye, Cui and
  Yu}]{wang2019heterogeneous}
\bibinfo{author}{Wang, X.}, \bibinfo{author}{Ji, H.}, \bibinfo{author}{Shi,
  C.}, \bibinfo{author}{Wang, B.}, \bibinfo{author}{Ye, Y.},
  \bibinfo{author}{Cui, P.}, \bibinfo{author}{Yu, P.S.}, \bibinfo{year}{2019}.
\newblock \bibinfo{title}{Heterogeneous graph attention network}, in:
  \bibinfo{booktitle}{The world wide web conference}, pp.
  \bibinfo{pages}{2022--2032}.
%Type = Inproceedings
\bibitem[{Wang et~al.(2018)Wang, Sun, Han, Liu and Zhu}]{wang2018sentiment}
\bibinfo{author}{Wang, Y.}, \bibinfo{author}{Sun, A.}, \bibinfo{author}{Han,
  J.}, \bibinfo{author}{Liu, Y.}, \bibinfo{author}{Zhu, X.},
  \bibinfo{year}{2018}.
\newblock \bibinfo{title}{Sentiment analysis by capsules}, in:
  \bibinfo{booktitle}{Proceedings of the 2018 world wide web conference}, pp.
  \bibinfo{pages}{1165--1174}.
%Type = Article
\bibitem[{Wang et~al.(2023)Wang, Wang, Zhan, Ma and Jiang}]{wang2023text}
\bibinfo{author}{Wang, Y.}, \bibinfo{author}{Wang, C.}, \bibinfo{author}{Zhan,
  J.}, \bibinfo{author}{Ma, W.}, \bibinfo{author}{Jiang, Y.},
  \bibinfo{year}{2023}.
\newblock \bibinfo{title}{Text fcg: Fusing contextual information via graph
  learning for text classification}.
\newblock \bibinfo{journal}{Expert Systems with Applications} ,
  \bibinfo{pages}{119658}.
%Type = Inproceedings
\bibitem[{Xia et~al.(2022)Xia, Wu, Chen, Hu and Li}]{xia2022simgrace}
\bibinfo{author}{Xia, J.}, \bibinfo{author}{Wu, L.}, \bibinfo{author}{Chen,
  J.}, \bibinfo{author}{Hu, B.}, \bibinfo{author}{Li, S.Z.},
  \bibinfo{year}{2022}.
\newblock \bibinfo{title}{Simgrace: A simple framework for graph contrastive
  learning without data augmentation}, in: \bibinfo{booktitle}{Proceedings of
  the ACM Web Conference 2022}, pp. \bibinfo{pages}{1070--1079}.
%Type = Article
\bibitem[{Xu et~al.(2021)Xu, Cheng, Luo, Chen and Zhang}]{xu2021infogcl}
\bibinfo{author}{Xu, D.}, \bibinfo{author}{Cheng, W.}, \bibinfo{author}{Luo,
  D.}, \bibinfo{author}{Chen, H.}, \bibinfo{author}{Zhang, X.},
  \bibinfo{year}{2021}.
\newblock \bibinfo{title}{Infogcl: Information-aware graph contrastive
  learning}.
\newblock \bibinfo{journal}{Advances in Neural Information Processing Systems}
  \bibinfo{volume}{34}, \bibinfo{pages}{30414--30425}.
%Type = Inproceedings
\bibitem[{Yang et~al.(2022a)Yang, Chen, Pan, Li, Yu and Xu}]{yang2022dual}
\bibinfo{author}{Yang, H.}, \bibinfo{author}{Chen, H.}, \bibinfo{author}{Pan,
  S.}, \bibinfo{author}{Li, L.}, \bibinfo{author}{Yu, P.S.},
  \bibinfo{author}{Xu, G.}, \bibinfo{year}{2022}a.
\newblock \bibinfo{title}{Dual space graph contrastive learning}, in:
  \bibinfo{booktitle}{Proceedings of the ACM Web Conference 2022}, pp.
  \bibinfo{pages}{1238--1247}.
%Type = Article
\bibitem[{Yang et~al.(2022b)Yang, Miao, Wang and Wang}]{yang2022contrastive}
\bibinfo{author}{Yang, Y.}, \bibinfo{author}{Miao, R.}, \bibinfo{author}{Wang,
  Y.}, \bibinfo{author}{Wang, X.}, \bibinfo{year}{2022}b.
\newblock \bibinfo{title}{Contrastive graph convolutional networks with
  adaptive augmentation for text classification}.
\newblock \bibinfo{journal}{Information Processing \& Management}
  \bibinfo{volume}{59}, \bibinfo{pages}{102946}.
%Type = Inproceedings
\bibitem[{Yao et~al.(2019)Yao, Mao and Luo}]{yao2019graph}
\bibinfo{author}{Yao, L.}, \bibinfo{author}{Mao, C.}, \bibinfo{author}{Luo,
  Y.}, \bibinfo{year}{2019}.
\newblock \bibinfo{title}{Graph convolutional networks for text
  classification}, in: \bibinfo{booktitle}{Proceedings of the AAAI conference
  on artificial intelligence}, pp. \bibinfo{pages}{7370--7377}.
%Type = Inproceedings
\bibitem[{Ying et~al.(2021)Ying, Shou and Liu}]{ying2021prediction}
\bibinfo{author}{Ying, R.}, \bibinfo{author}{Shou, Y.}, \bibinfo{author}{Liu,
  C.}, \bibinfo{year}{2021}.
\newblock \bibinfo{title}{Prediction model of dow jones index based on
  lstm-adaboost}, in: \bibinfo{booktitle}{2021 International Conference on
  Communications, Information System and Computer Engineering (CISCE)},
  \bibinfo{organization}{IEEE}. pp. \bibinfo{pages}{808--812}.
%Type = Article
\bibitem[{You et~al.(2020)You, Chen, Sui, Chen, Wang and Shen}]{you2020graph}
\bibinfo{author}{You, Y.}, \bibinfo{author}{Chen, T.}, \bibinfo{author}{Sui,
  Y.}, \bibinfo{author}{Chen, T.}, \bibinfo{author}{Wang, Z.},
  \bibinfo{author}{Shen, Y.}, \bibinfo{year}{2020}.
\newblock \bibinfo{title}{Graph contrastive learning with augmentations}.
\newblock \bibinfo{journal}{Advances in neural information processing systems}
  \bibinfo{volume}{33}, \bibinfo{pages}{5812--5823}.
%Type = Inproceedings
\bibitem[{Zhang and Zhang(2020)}]{zhang2020text}
\bibinfo{author}{Zhang, H.}, \bibinfo{author}{Zhang, J.}, \bibinfo{year}{2020}.
\newblock \bibinfo{title}{Text graph transformer for document classification},
  in: \bibinfo{booktitle}{Conference on empirical methods in natural language
  processing (EMNLP)}, pp. \bibinfo{pages}{1--9}.
%Type = Article
\bibitem[{Zhang et~al.(2022)Zhang, Deng, Ye, Zhang and Chen}]{zhang2022robust}
\bibinfo{author}{Zhang, N.}, \bibinfo{author}{Deng, S.}, \bibinfo{author}{Ye,
  H.}, \bibinfo{author}{Zhang, W.}, \bibinfo{author}{Chen, H.},
  \bibinfo{year}{2022}.
\newblock \bibinfo{title}{Robust triple extraction with cascade bidirectional
  capsule network}.
\newblock \bibinfo{journal}{Expert Systems with Applications}
  \bibinfo{volume}{187}, \bibinfo{pages}{115806}.
%Type = Article
\bibitem[{Zhang et~al.(2020)Zhang, Wang, Sun and Xiao}]{zhang2020practical}
\bibinfo{author}{Zhang, S.}, \bibinfo{author}{Wang, L.}, \bibinfo{author}{Sun,
  K.}, \bibinfo{author}{Xiao, X.}, \bibinfo{year}{2020}.
\newblock \bibinfo{title}{A practical chinese dependency parser based on a
  large-scale dataset}.
\newblock \bibinfo{journal}{arXiv preprint arXiv:2009.00901} .
%Type = Article
\bibitem[{Zhang et~al.(2024)Zhang, Shou, Meng, Ai and Li}]{zhang2024multi}
\bibinfo{author}{Zhang, Y.}, \bibinfo{author}{Shou, Y.}, \bibinfo{author}{Meng,
  T.}, \bibinfo{author}{Ai, W.}, \bibinfo{author}{Li, K.},
  \bibinfo{year}{2024}.
\newblock \bibinfo{title}{A multi-view mask contrastive learning graph
  convolutional neural network for age estimation}.
\newblock \bibinfo{journal}{Knowledge and Information Systems} ,
  \bibinfo{pages}{1--26}.
%Type = Article
\bibitem[{Zhang et~al.(2021)Zhang, Zhang, Qi, Manning and
  Langlotz}]{zhang2021biomedical}
\bibinfo{author}{Zhang, Y.}, \bibinfo{author}{Zhang, Y.}, \bibinfo{author}{Qi,
  P.}, \bibinfo{author}{Manning, C.D.}, \bibinfo{author}{Langlotz, C.P.},
  \bibinfo{year}{2021}.
\newblock \bibinfo{title}{Biomedical and clinical english model packages for
  the stanza python nlp library}.
\newblock \bibinfo{journal}{Journal of the American Medical Informatics
  Association} \bibinfo{volume}{28}, \bibinfo{pages}{1892--1899}.
%Type = Inproceedings
\bibitem[{Zhao and Song(2023)}]{zhao2023textgcl}
\bibinfo{author}{Zhao, Y.}, \bibinfo{author}{Song, X.}, \bibinfo{year}{2023}.
\newblock \bibinfo{title}{Textgcl: Graph contrastive learning for transductive
  text classification}, in: \bibinfo{booktitle}{2023 International Joint
  Conference on Neural Networks (IJCNN)}, \bibinfo{organization}{IEEE}. pp.
  \bibinfo{pages}{1--8}.
%Type = Inproceedings
\bibitem[{Zhu et~al.(2021)Zhu, Xu, Yu, Liu, Wu and Wang}]{zhu2021graph}
\bibinfo{author}{Zhu, Y.}, \bibinfo{author}{Xu, Y.}, \bibinfo{author}{Yu, F.},
  \bibinfo{author}{Liu, Q.}, \bibinfo{author}{Wu, S.}, \bibinfo{author}{Wang,
  L.}, \bibinfo{year}{2021}.
\newblock \bibinfo{title}{Graph contrastive learning with adaptive
  augmentation}, in: \bibinfo{booktitle}{Proceedings of the Web Conference
  2021}, pp. \bibinfo{pages}{2069--2080}.

\end{thebibliography}
	
	% % Biography
	% % Here goes the biography details.
	
	% % Here goes the biography details.
	
\end{document}